\documentclass{article}
\PassOptionsToPackage{table}{xcolor}
\usepackage{arxiv}
\usepackage[utf8]{inputenc}
\usepackage[T1]{fontenc}
\usepackage{natbib}
\setcitestyle{authoryear,round,citesep={;},aysep={,},yysep={;}}

\usepackage{amsmath,amsfonts,bm}
\usepackage{enumitem}
\def\eqref#1{equation~\ref{#1}}
\def\1{\bm{1}}

\DeclareMathAlphabet{\mathsfit}{\encodingdefault}{\sfdefault}{m}{sl}
\SetMathAlphabet{\mathsfit}{bold}{\encodingdefault}{\sfdefault}{bx}{n}

\usepackage{hyperref}
\usepackage{listings}
\usepackage{url}
\usepackage{graphicx}
\usepackage{float}
\usepackage{wrapfig}
\usepackage{booktabs}
\usepackage{array}
\usepackage{amssymb}
\usepackage{xcolor}
\definecolor{gazeanywheregray}{gray}{0.90}
\newcommand{\ours}{\rowcolor{gazeanywheregray}}
\newcommand{\losstick}{%
  \textcolor{green!45!black}{\ensuremath{\checkmark}}}
\newcommand{\losscross}{%
  \textcolor{red!80!black}{\ensuremath{\times}}}

\newcommand{\fitwidth}[1]{\resizebox{\ifdim\width>\linewidth\linewidth\else\width\fi}{!}{#1}}

\usepackage{xspace}
\usepackage{titletoc}
\usepackage{cleveref}   % after hyperref, as cleveref requires
\newcommand{\modelname}{GWM\xspace}
\title{Gaze World Model}
\renewcommand{\shorttitle}{Gaze World Model}
\date{}

\usepackage{authblk}

\author[1,*]{Jiawen Zhang}
\author[1,*]{Houze Yang}
\author[1]{Xu Cao}
\author[1]{Yifan Shen}
\author[2]{Zhongyi Zhou}
\author[1]{Steve Wong}
\author[1]{Ismini Lourentzou}
\author[1]{Inki Kim}
\author[1]{James M. Rehg}
\affil[1]{University of Illinois Urbana-Champaign}
\affil[2]{Google}
\affil[*]{Equal contribution}
\affil[ ]{Correspondence to: \texttt{\{xucao2,jrehg\}@illinois.edu}}

\begin{document}

\maketitle

\begin{figure}[!ht]
    \centering
    \vspace{-0.9cm}
    \includegraphics[width=0.99\textwidth]{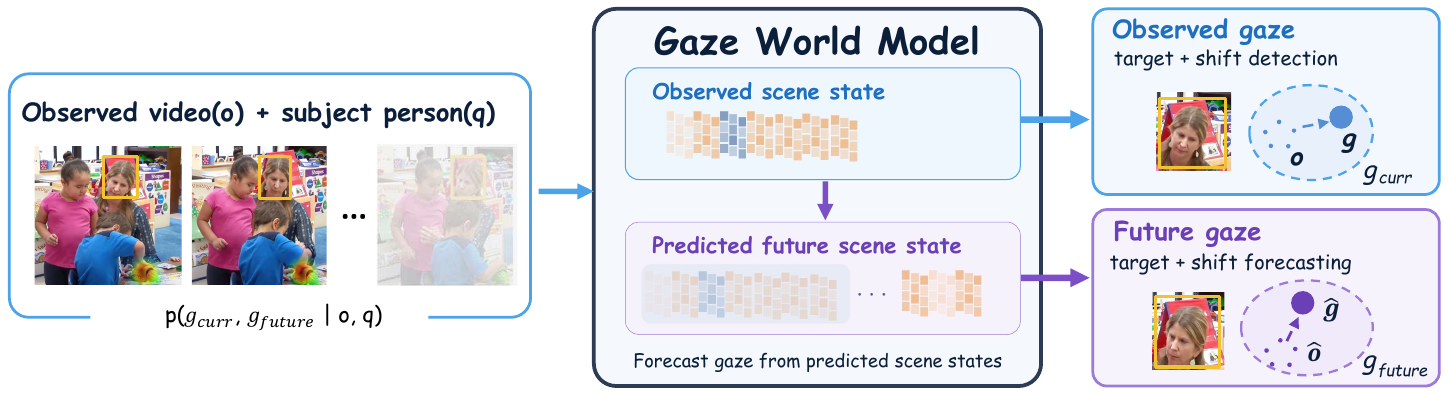}
    \caption{Overview of the Gaze World Model. Given the observation, the model estimates the current gaze of the specified person and rolls forward latent scene dynamics to forecast future gaze targets and shift events beyond the observation window.}
    \label{fig:overview}
    \vspace{0.3cm}
\end{figure}

\begin{abstract}
Where a person looks indicates their focus of attention, and how their gaze shifts between targets reveals how that attention responds to a changing scene.
Existing gaze target estimation methods largely operate frame by frame and do not model these shifts.
Gaze shifts are often driven by dynamic changes in the scene, which is the kind of information world models are designed to encode.
This relationship is subtle, however, since an event far across the frame may produce only a slight movement of the person's head or eyes.
We therefore introduce the Gaze World Model (GWM), which co-trains latent scene prediction with person-conditioned gaze supervision on a frozen V-JEPA~2.1 encoder; in our ablations, the latent prediction objective improves both shift detection and gaze localization.
Given a video clip and the person's head box, GWM estimates gaze locations and detects gaze shifts in the observed clip, and forecasts future gaze and shifts from predicted scene features.
To train and evaluate these tasks, we augment VideoAttentionTarget and ChildPlay with event-level gaze-shift annotations.
GWM achieves state-of-the-art shift detection on these benchmarks, about 10 points higher in four-class AP than the strongest baseline, while remaining comparable to the strongest image-based models in gaze target estimation, and outperforms the compared baselines on future gaze localization and shift forecasting.
\looseness=-1
\end{abstract}

% !TEX root = ../main.tex
\section{Introduction}
\label{sec:intro}

Social gaze, which includes mutual gaze, gaze following, and joint attention to a common object, is a key building block of social communication~\citep{emery2000eyes,tomasello1986joint}. In development, infants follow an adult's gaze within the first year of life, and episodes of joint attention support early word learning \citep{scaife1975capacity,tomasello1986joint}. For example,
during play a child may first look at an adult's face and then shift their gaze to a toy that the adult is holding \citep{land2006eyemovements,frischen2007gazecueing}.
The child's gaze location indicates their current focus of attention, and the shift from the adult to the toy can be part of a non-verbal joint-attention or requesting bid, through which the child communicates interest in the toy \citep{mundy1986defining}.
%%communicate indicates how this focus changes as the interaction progresses.
Thus, while gaze locations identify the visual inputs a child is receiving, the timing of the child's gaze shifts in the social context, which signal their intent and engagement, convey information that cannot be extracted from the gaze locations alone.
Automated coding of such social gaze in child--adult interaction is therefore of clinical interest \citep{chong2020eyecontact}, and a decline in eye looking from 2 to 6 months is among the earliest known markers of autism \citep{jones2013attention}.
%interest  shifts therefore reflect how a person perceives a changing scene and how their attention is reallocated over time, which gaze locations estimated at individual time points cannot capture.
\looseness=-1

Most prior works on gaze target estimation (GTE) have focused on localizing gaze within a single image \citep{recasens2015gazefollow,lian2018gaze,chong2018connecting,ryan2025gazelle,ye2026page}.
Video-based methods add temporal context through recurrent or temporal-attention modules over per-frame features \citep{chong2020vat,miao2023patch,gupta2024mtgs}, mainly to improve per-frame localization and in/out prediction, or locate the gaze target in other frames of the video \citep{recasens2017following}; video benchmarks such as VideoAttentionTarget and ChildPlay evaluate gaze frame by frame \citep{chong2020vat,tafasca2023childplay}.
These models output a gaze point in each frame and do not model or forecast transitions between gaze targets; gaze shifts appear only as a per-frame annotation label \citep{tafasca2023childplay} or as a separate downstream detector applied to per-frame attention maps \citep{chong2020vat}.
Social-gaze methods recognize mutual gaze, shared attention, and gaze-communication events such as aversion and following \citep{marinjimenez2019laeonet,fan2018sharedattention,fan2019gazecommunication,gupta2024mtgs}, but they label social-gaze states or communication events between people rather than detecting when a specified person's gaze moves from one target to another. The goal of this paper is to develop a model which is capable of \emph{explicitly predicting temporal shifts in gaze}, along with the evidence for those shifts in the form of changing gaze locations over time.
%The central question of this paper is how to infer these shifts.
\looseness=-1

A natural approach is to treat gaze-shift detection as an action recognition problem, and learn the temporal signature of a shift, such as a head turn~\citep{wang2025actiondetail}.
But a gaze shift is always accompanied by a change in the gaze target, and accurate inference therefore requires an understanding of the structure of the scene as well as the person.
Since gaze shifts are often driven by the events in a person's environment, such as the introduction of a novel object, an effective model for gaze shift detection must be attuned to the manner in which scenes change over time. Fortunately, recent works in world modeling have introduced a powerful approach 
%%To detect them, a model needs to understand how the scene changes over time, which is what world models are built to do.
to encode the current state of an environment and predict how it evolves over time~\citep{ha2018worldmodels,hafner2023dreamerv3}.
\looseness=-1

Pretrained world-model encoders such as V-JEPA \citep{assran2025vjepa2,murlabadia2026vjepa21} capture how a scene evolves within the observed video, but they are not trained to predict beyond it, nor to model how a particular person might respond to these changes.
The relationship between a scene change and the resulting gaze shift can be subtle, as a 
person's face or eyes may change only slightly in response to a distal event, whereas world models are primarily effective at capturing causal relationships that are readily visible in the video.
In order to overcome this challenge, we develop a procedure for co-training latent scene prediction with person-conditioned gaze supervision on top of a frozen V-JEPA encoder. This approach sets a new state of the art for gaze shift detection, and its gaze location estimates are competitive with those of the best image-based methods. To our knowledge, \modelname{} is the first model that both detects and forecasts typed transitions between semantic gaze targets of a specified person in unconstrained third-person video; prior work detects a single shift type from per-frame attention maps \citep{chong2020vat} or forecasts future gaze locations without modeling shifts \citep{kuang2026sage}.
\looseness=-1

Building on this insight, we propose the Gaze World Model (\modelname; Figure~\ref{fig:overview}).
Given an observed clip and a head box for the specified person, \modelname{} encodes the observed scene with a frozen V-JEPA~2.1 encoder \citep{murlabadia2026vjepa21}, conditions the scene features on the person's head through scene--head interaction, and aggregates them over time to estimate gaze and detect shifts.
A co-trained world predictor predicts future scene features, from which gaze decoders forecast future gaze and shifts.
At inference, \modelname{} requires only the observed video and the head box.
Our contributions are as follows:
\begin{itemize}[itemsep=0.5ex, parsep=0pt, topsep=-2.3pt, leftmargin=0.6cm]
    \item We formulate gaze modeling for a specified person as a world-modeling problem, covering gaze point estimation, gaze-shift detection, and forecasting over observed and future intervals.
    \item We propose \modelname, which co-trains latent world prediction with person-conditioned gaze supervision on a frozen V-JEPA~2.1 encoder and reads out gaze shifts from the learned representation; the latent objective improves shift detection and localization over gaze supervision alone.
    \item We introduce a gaze-shift dataset built on VideoAttentionTarget and ChildPlay, with event-level gaze-shift annotations over 4 hours of person-level gaze tracks, for training and evaluating gaze-shift detection and forecasting.
\end{itemize}

% !TEX root = ../main.tex
\section{Related Work}
\label{sec:related}

\noindent \textbf{Human Gaze Modeling.}
Gaze following predicts a spatial target distribution from a person's appearance and head orientation \citep{recasens2015gazefollow,lian2018gaze,chong2018connecting}. Later image methods add depth and 3D scene cues \citep{fang2021dual,bao2022escnet}, end-to-end head--target detection \citep{tu2022endtoend}, and object or semantic target reasoning \citep{wang2022gatector,tonini2023objectaware,tafasca2024semantic}. Video methods add temporal modeling and out-of-frame prediction \citep{chong2020vat,miao2023patch,gupta2024mtgs} or locate the gaze target in other frames of the video \citep{recasens2017following}, evaluated on video benchmarks such as VideoAttentionTarget and ChildPlay \citep{chong2020vat,tafasca2023childplay}, while recent spatial models build on large pretrained encoders, frozen \citep{ryan2025gazelle} or fine-tuned \citep{song2024vitgaze,ye2026page}. Subject conditioning identifies whose gaze to predict, through head boxes or textual and multimodal cues \citep{cao2026gazeanywhere,she2026textgaze,miao2026omnigf}. Gaze-event analysis and anticipation address a different temporal question: classical and learned methods identify fixations and saccades from ocular signals \citep{salvucci2000identifying,holmqvist2011eye,hessels2024fundamentals1,zemblys2019gazenet,startsev2019cnnblstm}; egocentric methods estimate gaze online or jointly with actions \citep{li2018eyebeholder,li2026argaze} and exploit learned attention transitions between fixations \citep{huang2018attentiontransition}; and anticipation methods predict future gaze or behavior \citep{zhang2017dfg,furnari2020rulstm,kuang2026sage}. We study target-transition events for a specified person, with detection in observed video and forecasting beyond it. Appendix~\ref{app:extended-related-work} provides the extended discussion. \looseness=-1

\noindent \textbf{World Models and Behavioral Prediction.}
World models learn transition dynamics that support prediction and planning by rolling the current state forward under possible actions \citep{ha2018worldmodels,hafner2023dreamerv3,ding2024worldmodelsurvey}. Latent world models instead predict future patch features of a frozen pretrained encoder, conditioned on agent actions \citep{zhou2025dinowm,assran2025vjepa2}; predicting future visual representations likewise supports action anticipation \citep{vondrick2016anticipating,girdhar2021avt}. In third-person video, person-level forecasting has targeted future paths and activities \citep{liang2019peeking}, and a broader line of work forecasts non-verbal social behavior in interactions \citep{barquero2022didnt}. Behavioral world modeling shifts the emphasis from predicting scene appearance alone to predicting how agents act and how their interactions change the world \citep{shen2026wam}. Existing approaches learn structured human-centric actions and affordances from human videos \citep{mendonca2023structured}, or jointly predict an egocentric agent's future action and world state \citep{chen2025egoagent}; related gaze--action models likewise use attention as a cue to future human behavior \citep{huang2015gazeintent,kuang2026sage}. These models primarily address embodied control or camera-wearer actions. We model a specified person's semantic gaze target in third-person video as a compact behavioral state: target changes define observed events, and world prediction supports future gaze and shift forecasting. \looseness=-1
% \il{need a bit of ending in some place with how our work differs or what our work introduces.}\il{unclear if the last sentence is needed here, i.e., about discriminative backbones.}

% !TEX root = ../main.tex
\section{Gaze World Model}
\label{sec:method}

The Gaze World Model (GWM) estimates and forecasts gaze for a specified person.
We distinguish annotated events, benchmark labels, and model outputs, then describe
how subject conditioning, temporal aggregation, and latent prediction support these tasks.

\subsection{Task Definition}
\label{sec:GWM-task}

\begin{figure}[t]
    \centering
    \includegraphics[width=0.99\linewidth]{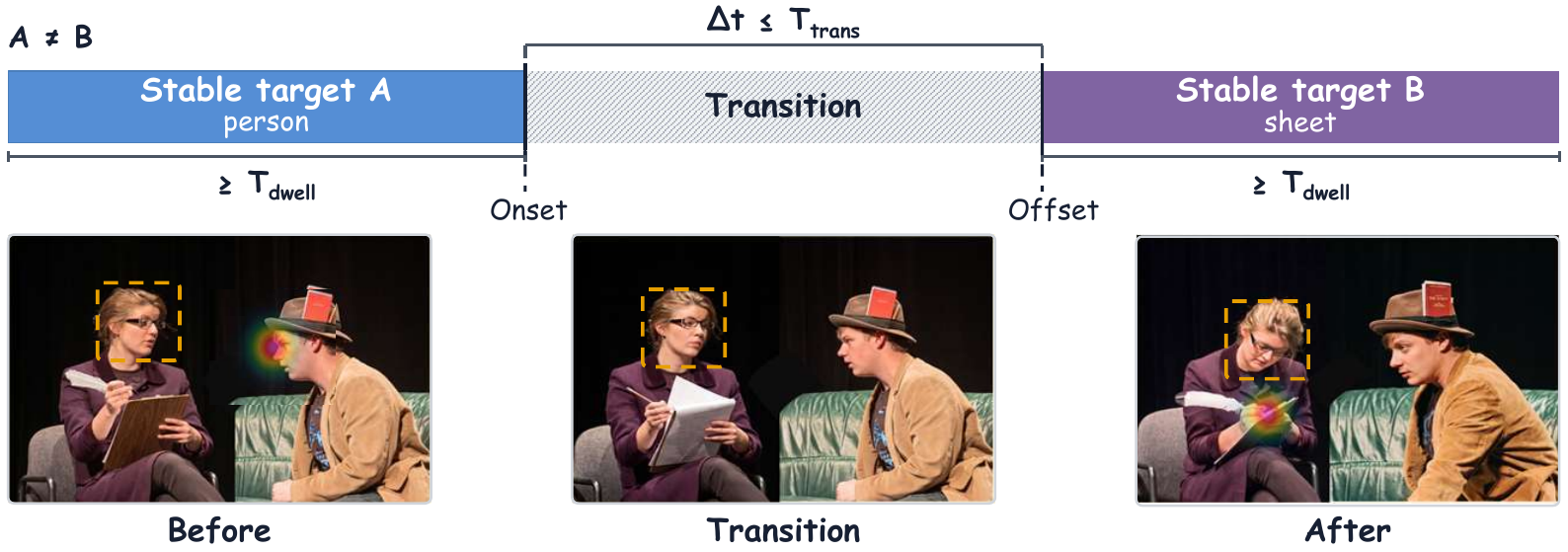}
    % \vspace{-0.1cm}
    \caption{Definition of a gaze-shift event. Per-frame target labels of the specified person (dashed box) show a stable fixation on $A$, a transition interval, and a stable fixation on $B$.}
    \label{fig:gaze-shift}
    % \vspace{-0.2cm}
\end{figure}

In Figure~\ref{fig:gaze-shift}, the specified person shifts gaze from another
person ($A$) to a sheet ($B$), illustrating a human-to-other transition.
A gaze shift consists of stable fixations on two different targets ($A \neq B$),
each lasting at least $T_{\text{dwell}}$, separated by a transition interval of at most
$T_{\text{trans}}$ (Appendix Table~\ref{tab:rule-thresholds}).
We define three person-involving transition types as positive:
human-to-human between distinct people, human-to-other, and other-to-human.
Here, \texttt{other} includes objects, background regions, and out-of-frame targets.
Changes between body parts of the same person or between \texttt{other}
targets are negative.
Alongside gaze-point estimates, detection assigns each observed window one of the three
transition labels or no in-scope shift. Forecasting predicts whether a transition of
any of the same three types begins in the future window.
Neither task predicts exact event boundaries or target instance identifiers.
Both tasks use the cleaned annotations produced by our data pipeline.
Appendix Table~\ref{tab:gaze-task-terms} summarizes the terminology and provides
the window-label assignment rule. \looseness=-1

% Keep the architecture figure off the task-definition page.
\suppressfloats[t]
\begin{figure}[t]
    \centering
    \includegraphics[width=0.95\linewidth]{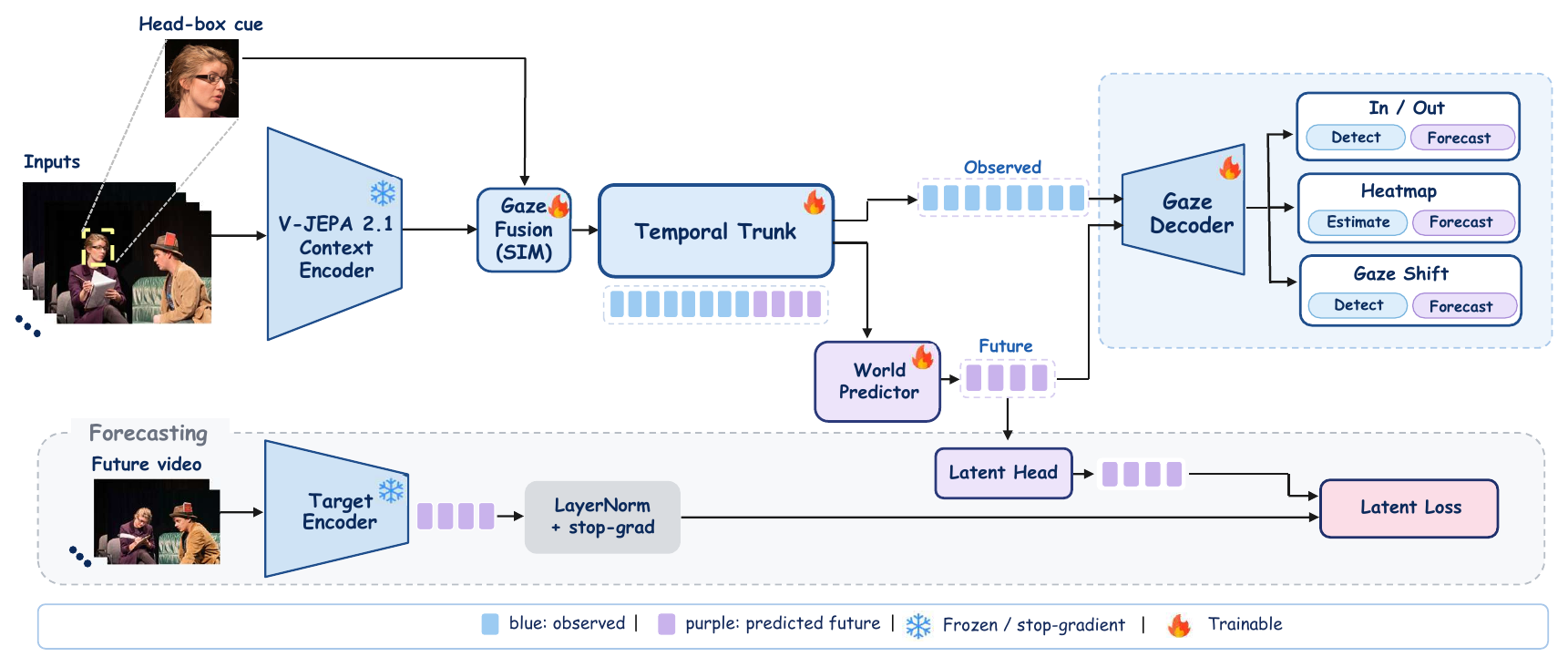}
    \caption{Architecture of \modelname. The frozen V-JEPA~2.1 encoder encodes the observed clip, and SIM conditions the scene features on the head box. The temporal trunk processes observed and future frame tokens, and the world predictor refines the future positions. Task decoders read the observed positions for estimation and detection, and the predicted future positions for forecasting. }     \label{fig:GWM-architecture}
    \vspace{-2mm}
\end{figure}

\subsection{Model Architecture}
\label{sec:architecture}

Figure~\ref{fig:GWM-architecture} summarizes one architecture with a shared frozen visual encoder, scene--head interaction (SIM), a temporal trunk, and a world-prediction branch. Three groups of task heads provide gaze heatmaps and in/out predictions, gaze-shift detection, and gaze-shift forecasting. \looseness=-1

\noindent \textbf{Scene Representation Encoder.}
A frozen V-JEPA~2.1 encoder \citep{murlabadia2026vjepa21} encodes the observed scene frames. In the main forecasting setting, \modelname observes 16 frames and forecasts gaze over the next 8 frames at 8 fps. Each tubelet spans two frames, giving $T_o=8$ observed and $T_f=4$ future positions, and each scene tubelet yields $P=32\times32=1024$ patch features of width $C$. Future frames are never encoded on the prediction path; they are used only to form latent targets during video training. \looseness=-1

% \noindent \textbf{Subject-Conditioned Fusion.} Separate linear projections map scene and head crops to width $D=256$. SIM conditions the scene features on the specified person through interaction between a scene stream and a head stream. The head stream is built from the person's head crops, giving $P_h=16\times16=256$ patch features per tubelet (Appendix~\ref{app:model-details}). Each stream receives $R=5$ prefix tokens: one in/out token at position zero and four register tokens. Thus, each scene tubelet contains $N=P+R=1029$ tokens, and each head tubelet contains $P_h+R=261$ tokens. The head box for a tubelet is the mean of its two frame-level boxes. We map this box to a rectangle in scene-grid coordinates and assign head-patch RoPE coordinates within that rectangle. Each stream first undergoes self-attention, followed by five SIM layers with bidirectional scene--head cross-attention and stream-specific self-attention. The box-to-coordinate mapping is parameter-free; the projections and attention layers are learned. During video training, the pretrained projections and SIM are frozen and evaluated without gradient tracking. \looseness=-1

\noindent \textbf{Subject-Conditioned Fusion.}
The scene--head interaction module (SIM) of PaGE~\citep{ye2026page} conditions the scene features on the specified person. It pairs the scene stream with a head stream from the person's head crops (Appendix~\ref{app:model-details}), and linear projections map both streams to width $D=256$. Each stream is prefixed with $R=5$ tokens, an in/out token at position zero and four register tokens, so a scene tubelet has $N=P+R=1029$ tokens. The head box, averaged over the two frames of a tubelet, places the head tokens in the scene by assigning them RoPE coordinates inside the box's scene-grid rectangle. After one self-attention layer per stream, five SIM layers each apply bidirectional scene--head cross-attention and stream-specific self-attention \citep{vaswani2017attention}. SIM is trained in the image stage and kept frozen during video training. \looseness=-1

\noindent \textbf{Temporal Trunk.}
The trunk aggregates person-conditioned scene features across tubelets using joint self-attention with three-dimensional RoPE coordinates $(t,y,x)$ and SwiGLU MLPs. Each scene position retains its $N$ tokens. For forecasting, a normalized linear projection of the raw observed encoder features is added to the corresponding SIM patch tokens, with zero additions at prefix positions. Four future query bundles are initialized by copying the last observed bundle (index $7$) and adding a learned future embedding. The two trunk blocks process all $T_o+T_f=12$ positions. Observed queries attend only to observed positions; future queries attend to both observed and future positions. Thus, attention masking controls access among token positions, while the future queries themselves contain no future pixels. The head-token stream bypasses both the trunk and the world branch. \looseness=-1

\noindent \textbf{World Predictor.}
The world branch applies two additional RoPE transformer blocks to all 12 trunk positions with the same attention mask. Observed gaze predictions read the first eight trunk outputs, while future gaze predictions read the last four world-branch outputs. Future positions originate from the copied observation and future embedding.%, not from replacing an encoded future clip with mask tokens.\looseness=-1

\begingroup
\setlength{\abovedisplayskip}{4pt}
\setlength{\belowdisplayskip}{4pt}
\setlength{\abovedisplayshortskip}{0pt}
\setlength{\belowdisplayshortskip}{4pt}
\setlength{\jot}{1pt}
During video training, we supervise future patch representations by latent prediction \citep{assran2023ijepa,bardes2024vjepa}, here of future frames against the frozen encoder, as in \citep{assran2025vjepa2,zhou2025dinowm}. During training only, the same frozen encoder separately processes the eight real future scene frames without gradient tracking. Its patch features are layer-normalized and detached to form targets $z_{t,p}$, where $t\in\{8,\ldots,11\}$ and $p\in\{1,\ldots,P\}$. Let $e_{s,p}$ denote the raw observed encoder features and $u_{t,p}$ the world-branch patch outputs. The latent prediction is a residual on the normalized observation mean,
\[
\hat z_{t,p}=A\,\mathrm{LN}(u_{t,p})+b+
\mathrm{LN}\!\left(\frac{1}{T_o}\sum_{s=0}^{T_o-1}e_{s,p}\right),
\]
where $A:\mathbb{R}^{D}\rightarrow\mathbb{R}^{C}$ and $b$ are zero-initialized. The motion-weighted objective is
\begin{equation}
\label{eq:world-prediction}
\mathcal{L}_{\mathrm{lat}}=
\frac{1}{|\mathcal M|C}\sum_{(t,p)\in\mathcal M}
w_{t,p}\,\lVert\hat z_{t,p}-z_{t,p}\rVert_1.
\end{equation}
\endgroup
  The weights $w_{t,p}$ emphasize patches with larger
  temporal feature changes and are detached from gradient
  computation. 
  % The normalized change is the mean absolute difference between $z_{t,p}$ and the last observed feature $\mathrm{LN}(e_{7,p})$, divided by its mean across patches for the same sample and future time step (with the denominator clamped to at least $10^{-6}$). The weights are detached from gradient computation. 
Here $\mathcal M$ contains the $T_f\times P$ future patch positions per sample, excluding all prefix tokens, and the L1 norm sums over the $C$ feature channels. The loss is also averaged over the batch and contributes $\lambda_{\mathrm{lat}}\mathcal{L}_{\mathrm{lat}}$ with $\lambda_{\mathrm{lat}}=0.1$. Targets come from the shared frozen encoder; there is no EMA encoder or target predictor. Separate observed/future encoding keeps future pixels outside the prediction path, including during training. Through the shared trunk, $\mathcal{L}_{\mathrm{lat}}$ also shapes the features used for observed gaze and detection. \looseness=-1

\noindent \textbf{Task Decoders.}
\label{sec:decoders}
Let $G_{t'}$ denote the scene tokens selected from the trunk for observed positions or from the world branch for future positions, and let $H_{t'}$ denote the observed SIM head tokens.
\begin{itemize}[leftmargin=1.2em,topsep=1pt,itemsep=1pt,parsep=0pt]
    \item \textbf{Gaze heatmap and in/out heads.}
    A shared heatmap head upsamples the scene patch grid with a transposed convolution
    and a $1\times1$ convolution to obtain $64\times64$ heatmap logits $\hat M_{t'}$.
    For observed positions, the in/out head concatenates the position-zero scene and head tokens
    and applies an MLP to obtain the logit $\hat v_{t'}$; its sigmoid gives the in-frame probability.
    Future positions reuse the last observed head in/out token, $H_{7,0}$.
    Both $\hat M$ and $\hat v$ are logits; training uses binary cross-entropy with logits on valid labels. Each tubelet prediction is repeated over its two constituent frames to align with frame-level supervision. \looseness=-1

    \item \textbf{Gaze-shift detection head.}
    Each observed tubelet is summarized by
    $x_{t'}=W_s[\bar g_{t'};q_{t'};m_{t'};h_{t'}]+b_s$.
    Here $\bar g_{t'}=R^{-1}\sum_{r=0}^{R-1}G_{t',r}$ averages all five scene prefix tokens,
    $q_{t'}$ pools scene patches with heatmap-derived attention,
    $m_{t'}$ is an MLP embedding of seven scalars (soft-argmax $x,y$, heatmap peak, in/out probability, $\Delta x,\Delta y$, and displacement magnitude), and
    $h_{t'}$ averages all head-stream tokens.
    A transformer layer aggregates the eight summary tokens with temporal embeddings and a learned class token, followed by a linear classifier with $1+3$ logits for the negative class and the three positive transition types. \looseness=-1

    \item \textbf{Gaze-shift forecasting head.}
    A separate head uses the same four-part summary design and temporal aggregation, but has two output logits for an in-scope shift versus no in-scope shift in the future interval. It reads all 12 tubelets: eight observed trunk outputs and four future world-branch outputs. For observed tubelets, $h_{t'}$ is the head-token mean; for future tubelets, it is a learned placeholder vector. The forecasting head is trained with two-class cross-entropy.
\end{itemize}

% !TEX root = ../main.tex
\section{Gaze-Shift Dataset}
\label{sec:pipeline}

Frame-level gaze coordinates are insufficient to supervise gaze shifts: they do not indicate
whether neighboring coordinates refer to the same target instance, where a transition begins and
ends, or which targets it connects. Exhaustively annotating these events in unconstrained video is
costly and prone to disagreement between annotators. We therefore convert the frame-level
annotations in VideoAttentionTarget (VAT) \citep{chong2020vat} and ChildPlay
\citep{tafasca2023childplay} into event-level transitions and window-level training labels, using a
semi-automated engine that proposes candidate transitions from spatial and temporal rules, verifies
them with a multimodal model, and submits sampled candidates to iterative human verification.
Appendix~\ref{app:data-engine} specifies the engine in full.
Human targets are distinguished by person identifiers propagated from the rule stage.
Held-object keys encode an association with a person, while independent objects use semantic
keys; neither representation guarantees persistent object-instance identity. \looseness=-1

We assign window labels using the person-involving gaze transitions defined in
Section~\ref{sec:GWM-task}. This scope captures changes in interaction partner, speaker, or
social focus while distinguishing \texttt{human} endpoints from \texttt{other} targets.
Transitions between body parts of the same person, such as head-to-hand
(\texttt{same\_person}), and transitions between \texttt{other} targets
(\texttt{other2other}) do not constitute an in-scope shift, even when gaze visibly moves.
A no in-scope shift label therefore indicates the absence of the three benchmark transition
types, not necessarily a stationary gaze point. Cases with unresolved event status or
required annotations are excluded, rather than treated as negatives
(Appendix~\ref{app:semantic-verification}). \looseness=-1

The source videos run at 25 to 30 fps, and event construction uses the native frame
rate of each clip. After annotation, both datasets are sampled at 8 fps and extracted with a
sliding window of 16 frames, so each model input covers 2 seconds. We split the data by source video.
We use the official ChildPlay train, validation, and test splits. VideoAttentionTarget has no
official validation split, so we hold out a fixed 10\% of the training shows as validation.
Hyperparameters, including $\lambda_{\mathrm{lat}}$, the class weighting, and the temporal
sampling, and all checkpoints are selected on the validation splits; the test splits are used
only to report results.
The annotation set contains 6,204 confirmed transitions, and training uses 15,086 windows. \looseness=-1

% !TEX root = ../main.tex
\section{Experiments}
\label{sec:exp}

We evaluate \modelname on the two tasks defined in Section~\ref{sec:GWM-task}: detection
within the observed clip and forecasting beyond it. Each task has two outputs for the specified
person, the gaze points and the gaze shifts. We compare against four groups of methods:
image-based GTE models that condition a pretrained encoder on the head box, a temporal video
gaze model, a video gaze-anticipation baseline, and general
multimodal models that receive the subject as a head box drawn on every frame.
Section~\ref{sec:main-results} first reports GTE and shift detection within the observed clip, then
the temporal consistency of the predicted gaze trajectories, and finally forecasting beyond the clip. Section~\ref{sec:ablations} examines the
temporal trunk, the latent prediction objective and its weight, the contribution of each loss
term used for video training, and freezing the visual encoder; analyses of the class weighting in
the shift detection loss and of the temporal sampling rate are also reported in
Appendix~\ref{app:supp-experiments}. \looseness=-1

\subsection{Implementation Details}
\label{sec:implementation}

\paragraph{Baselines.} We compare against Gaze-LLE \citep{ryan2025gazelle} and PaGE
\citep{ye2026page}, each at two encoder sizes; PaGE ViT-B Distill is distilled from the ViT-H+ model on a larger unlabeled image set. We also evaluate MTGS~\citep{gupta2024mtgs}, a temporal model that also predicts social gaze, with its released checkpoint. For GTE, we evaluate the released models of
Gaze-LLE, PaGE, and MTGS with ground-truth head boxes on the same test windows as \modelname, sampled at 8\,fps from the 25 to 30\,fps source videos. By contrast, the results reported in~\citep{ryan2025gazelle,ye2026page} were obtained on annotated test frames at the native frame rate, so our numbers differ from them. On ChildPlay, we evaluate Gaze-LLE with its released VideoAttentionTarget checkpoint. For shift detection, we keep each
gaze model frozen and train a two-layer temporal classifier on its per-frame features, in the same
way as the shift detection head of \modelname is trained on frozen upstream features.
\modelname thus models temporal context before the gaze readout, whereas the image baselines aggregate time only in the added classifier. The image-model and \modelname shift
classifiers use the same training windows and test clips. For binary evaluation, a prediction
in any of the three positive classes counts as shift; all others count as no shift.
Two further shift detectors use no gaze readout. The first trains the same temporal classifier on frozen per-frame V-JEPA~2.1 ViT-L features and the head box, without gaze supervision, so only the classifier is learned. The second applies the Stage-1 rules of Appendix~\ref{app:data-engine}, with the same SAM~3 masks and thresholds, to the per-frame predictions of PaGE ViT-H+ in place of the annotated gaze points, and involves no training; its MTGS variant also uses MTGS looking-at-humans (LAH) outputs (Appendix~\ref{app:baseline-adaptation}).
For forecasting, we compare against SAGE~\citep{kuang2026sage}, which has no public code; we re-implement it with two gaze detectors: Sharingan~\citep{tafasca2024sharingan}, the strongest exocentric detector evaluated in SAGE, and PaGE, the state-of-the-art image GTE model.

We also evaluate Qwen3.8~27B~\citep{qwenteam2026qwen38} and
Gemini~3.8~Flash~\citep{geminiteam2026gemini38} zero-shot on all tasks, marking the
specified person with a red head box on every frame and prompting for probabilities, which
serve as ranking scores for AP. \looseness=-1

\paragraph{Training Schedule.}
\label{sec:training-protocol}
\modelname is a single model trained in two stages. In the image stage,
we train subject conditioning and gaze readout on single frames, mainly from
GazeFollow~\citep{recasens2015gazefollow}, together with frames from the training splits of
VideoAttentionTarget and ChildPlay, and MVGT~\citep{miao2025multiview}.
In the video stage, the input projections and SIM are kept frozen. We first train the temporal
trunk, the world predictor, and the gaze heads jointly with the observed-gaze (heatmap and in/out),
future-gaze, and latent-prediction losses. We then freeze these components and train the
gaze-shift detection head and the binary forecasting head on their outputs. All reported results come from this model. \looseness=-1

We use V-JEPA~2.1 ViT-B/L/g encoders \citep{dosovitskiy2021vit,zhai2022scaling}. By default, the visual encoder is kept frozen in all stages;
Table~\ref{tab:ablation-unfreeze} examines unfreezing its last four blocks. Each input clip
contains 16 observed frames sampled at 8 fps (2\,s), and in the main forecasting setting the model
forecasts gaze for the following eight frames (1\,s). The image stage is trained on $4\times$
NVIDIA H100 GPUs. The first step of the video stage trains for 20 epochs at an effective batch size of 32 with AdamW, a peak learning rate of $5\times10^{-4}$, a weight decay of $0.05$, a linear warmup over 300 updates, and cosine decay to $1\times10^{-7}$. The shift heads train for 25 epochs at batch size 64 with the same schedule, a peak learning rate of $3\times10^{-4}$, and a warmup over 200 updates. \looseness=-1

\subsection{Main results}
\label{sec:main-results}

\begin{table}[t!]
\centering
\caption{Observed gaze estimation and shift detection. GTE reports heatmap AUC, normalized L2, and in/out AP; VLMs output points only, so AUC is not reported (--); V-JEPA has no gaze output. Event scores (\%) are binary AP/F1 and four-class macro-AP/macro-F1. Gaze and V-JEPA baselines detect shifts with a temporal classifier on frozen features; ``+ rules'' applies the Stage-1 rules (Appendix~\ref{app:data-engine}) to the model's gaze predictions, outputs hard labels (no AP), and repeats that model's Params (plus SAM~3 for masks) and GTE in \textcolor{gray}{gray}, unranked. Params count all inference parameters (PaGE has separate scene and head encoders). In all tables, best results are in \textbf{bold} and second best \underline{underlined}.}
\label{tab:main-gaze-point}
%\begingroup
\setlength{\tabcolsep}{4pt}
\renewcommand{\arraystretch}{1.0}
\definecolor{gazeours}{gray}{0.90}
\definecolor{gazegroup}{RGB}{80,88,98}
\resizebox{0.87\linewidth}{!}{%
\begin{tabular}{@{}l*{11}{c}@{}}

\toprule
& & \multicolumn{6}{c}{\textbf{Gaze-target estimation}}
& \multicolumn{4}{c}{\textbf{Gaze-shift detection}} \\
\cmidrule(lr){3-8}\cmidrule(lr){9-12}
\textbf{Model} & {\footnotesize\textbf{Params}}
& \multicolumn{3}{c}{\footnotesize\textbf{VideoAttentionTarget}}
& \multicolumn{3}{c}{\footnotesize\textbf{ChildPlay}}
& \multicolumn{2}{c}{\footnotesize\textbf{Binary}}
& \multicolumn{2}{c}{\footnotesize\textbf{Four-class}} \\
\cmidrule(lr){3-5}\cmidrule(lr){6-8}
\cmidrule(lr){9-10}\cmidrule(lr){11-12}
& & {\footnotesize AUC$\uparrow$} & {\footnotesize L2$\downarrow$} & {\footnotesize AP$\uparrow$}
& {\footnotesize AUC$\uparrow$} & {\footnotesize L2$\downarrow$} & {\footnotesize AP$\uparrow$}
& {\footnotesize AP $\uparrow$} & {\footnotesize F1$\uparrow$}
& {\footnotesize AP $\uparrow$} & {\footnotesize F1$\uparrow$} \\
\midrule
\multicolumn{12}{@{}l}{\itshape Image gaze models} \\
\addlinespace[1pt]
Gaze-LLE-B~\citep{ryan2025gazelle} & 87M & 0.945 & 0.089 & 0.885 & 0.953 & 0.108 & \underline{0.998} & 68.46 & 62.24 & 51.31 & 49.59 \\
Gaze-LLE-L~\citep{ryan2025gazelle} & 304M & 0.951 & 0.085 & 0.889 & 0.953 & 0.099 & 0.996 & 72.51 & 61.82 & 57.38 & 54.10\\
PaGE-B (distilled)~\citep{ye2026page} & 171M & 0.971  & 0.063 & \textbf{0.946} & 0.969 & 0.072 & 0.995 & 76.91 & 74.07 & 63.08 & 60.16 \\
PaGE-H+~\citep{ye2026page} & 1.7B & 0.971 & 0.065 & 0.943 & \underline{0.970} & \textbf{0.068} & 0.997 & 78.38 & 74.42 & 66.32 & 65.67 \\
\midrule
\multicolumn{12}{@{}l}{\itshape Video gaze models} \\
\addlinespace[1pt]
MTGS~\citep{gupta2024mtgs} & 157M & 0.924 & 0.080 & 0.882 & 0.905 & 0.112 & 0.996 & 72.82 & 67.85 & 51.47 & 57.40 \\
\midrule
\multicolumn{12}{@{}l}{\itshape Vision-language models} \\
\addlinespace[1pt]
Qwen3.8-27B~\citep{qwenteam2026qwen38} & 27B & -- & 0.155 & 0.789 & -- & 0.134 & 0.981 & 62.74 & 52.88 & 44.35& 49.47 \\
Gemini 3.8 Flash~\citep{geminiteam2026gemini38} & -- & -- & 0.123 & 0.828 & -- & 0.107 & 0.984 & 75.41 & 56.85 & 55.97 & 53.89 \\
\midrule
\multicolumn{12}{@{}l}{\itshape Non-gaze and rule-based detectors} \\
\addlinespace[1pt]
V-JEPA 2.1-L~\citep{murlabadia2026vjepa21} & 304M & \multicolumn{6}{c}{\textcolor{gray}{\textit{no gaze output}}} & 69.75 & 62.38 & 60.57 & 46.39 \\
PaGE-H+ + rules & \textcolor{gray}{1.7B} & \textcolor{gray}{0.971} & \textcolor{gray}{0.065} & \textcolor{gray}{0.943} & \textcolor{gray}{0.970} & \textcolor{gray}{0.068} & \textcolor{gray}{0.997} & -- & 64.94 & -- & 44.65 \\
MTGS + rules (LAH) & \textcolor{gray}{157M} & \textcolor{gray}{0.924} & \textcolor{gray}{0.080} & \textcolor{gray}{0.882} & \textcolor{gray}{0.905} & \textcolor{gray}{0.112} & \textcolor{gray}{0.996} & -- & 62.87 & -- & 47.45 \\
\midrule
\rowcolor{gazeours}
GWM-B (Ours) & 105M & 0.966 & 0.063 & 0.942 & 0.964 & 0.077 & 0.994 & 85.34 & 78.50 & 72.45 & 71.80 \\
\rowcolor{gazeours}
GWM-L (Ours) & 320M & \underline{0.973} & \underline{0.061} & \underline{0.945} & 0.968 & 0.073 & 0.997 & \underline{88.65} & \underline{82.17} & \textbf{76.09} & \textbf{75.05} \\
\rowcolor{gazeours}
GWM-g (Ours) & 1B & \textbf{0.974} & \textbf{0.060} & \underline{0.945} & \textbf{0.972} & \underline{0.069} & \textbf{0.999} & \textbf{89.72} & \textbf{82.96} & \underline{75.91} & \underline{74.87}   \\
\bottomrule
\end{tabular}
}
%\parbox{\linewidth}{\footnotesize
% \endgroup
\end{table}

\begin{table}[t]
\centering
\caption{Temporal consistency of predicted gaze trajectories.}
\label{tab:ablation-prediction-temporal}
\setlength{\tabcolsep}{3.5pt}
\resizebox{0.7\linewidth}{!}{%
\begin{tabular}{@{}lcccccc@{}}
\toprule
& \multicolumn{3}{c}{\textbf{VideoAttentionTarget}}
& \multicolumn{3}{c}{\textbf{ChildPlay}} \\
\cmidrule(lr){2-4}\cmidrule(lr){5-7}
\textbf{Model}
& \textbf{Jitter $\downarrow$} & \textbf{Flicker $\downarrow$} & \textbf{$r$ $\downarrow$}
& \textbf{Jitter $\downarrow$} & \textbf{Flicker $\downarrow$} & \textbf{$r$ $\downarrow$} \\
\midrule
Gaze-LLE-B~\citep{ryan2025gazelle} & 0.026 & 0.053 & 0.068 & 0.028 & 0.061 & 0.085 \\
Gaze-LLE-L~\citep{ryan2025gazelle} & 0.020 & 0.035 & 0.065 & 0.029 & 0.059 & 0.070 \\
PaGE-B (distilled)~\citep{ye2026page} & 0.022 & 0.048 & 0.048 & 0.026 & 0.063 & 0.058 \\
PaGE-H+~\citep{ye2026page} & 0.015 & 0.030 & 0.054 & 0.020 & 0.043 & 0.060 \\
\midrule
$-$ temporal trunk & 0.019 & 0.037 & 0.050 & 0.020 & 0.048 & 0.057 \\
\ours GWM-B (Ours) & \textbf{0.012} & 0.022 & 0.051 & 0.016 & \underline{0.030} & 0.049 \\
\ours GWM-L (Ours)
& \textbf{0.012} & \underline{0.020} & \textbf{0.044}
& \textbf{0.013} & \textbf{0.027} & \underline{0.048} \\
\ours GWM-g (Ours) & \underline{0.013} & \textbf{0.019} & \underline{0.047} & \underline{0.014} & \textbf{0.027} & \textbf{0.045} \\
\bottomrule
\end{tabular}
}
\end{table}

\begin{table}[t]
\centering
\caption{Gaze forecasting beyond the observation window. Avg. denotes the frame-wise mean L2 error over the future interval. Persistence repeats the last observed estimate. The future-shift positive rate is 45.12\% (random-score AP).}
\label{tab:main-forecast}
\footnotesize
\setlength{\tabcolsep}{4pt}
\begin{tabular}{@{}l*{9}{c}@{}}
\toprule
\textbf{Gaze Model}
& \multicolumn{6}{c}{\textbf{Future gaze point (L2 $\downarrow$)}}
& \multicolumn{3}{c}{\textbf{Future gaze shift}} \\
\cmidrule(lr){2-7} \cmidrule(lr){8-10}
& \multicolumn{3}{c}{\textbf{VideoAttentionTarget}}
& \multicolumn{3}{c}{\textbf{ChildPlay}}
& \multicolumn{3}{c}{\textbf{Binary}} \\
\cmidrule(lr){2-4} \cmidrule(lr){5-7} \cmidrule(lr){8-10}
& \textbf{0.5 s} & \textbf{1.0 s} & \textbf{Avg.}
& \textbf{0.5 s} & \textbf{1.0 s} & \textbf{Avg.}
& \textbf{AP $\uparrow$} & \textbf{F1 $\uparrow$} & \textbf{Recall $\uparrow$} \\
\midrule
SAGE~\citep{kuang2026sage} w/ Sharingan
& 0.137 & 0.167 & 0.142
& 0.162 & 0.181 & 0.166
& 56.51 &  65.65 & 78.30\\
SAGE~\citep{kuang2026sage} w/ PaGE
& 0.116 & 0.148 &  0.123
& 0.143 & 0.165 & 0.145
& 66.23&  66.67 & 71.40\\
Qwen3.8-27B~\citep{qwenteam2026qwen38}
& 0.189 & 0.215 & 0.193
& 0.177 & 0.178 & 0.177
& 55.47 & 41.18 & 49.79 \\
Gemini 3.8 Flash~\citep{geminiteam2026gemini38}
& 0.168 & 0.210 & 0.180
& 0.152 & 0.167 & 0.155
& 62.00 & 44.20 & 53.00 \\
\midrule
\multicolumn{10}{l}{\textit{Reference predictor}} \\
Persistence (GWM-L)
& 0.120 & 0.158 & 0.128
& 0.137 & 0.168 & 0.143
& -- & -- & -- \\
\midrule
\ours GWM-B (Ours)
& 0.117 & 0.145 & 0.120
& \underline{0.125} & 0.146 & 0.128
& 71.30 & 71.20 & 84.20 \\
\ours GWM-L (Ours)
& \underline{0.102} & \underline{0.132} & \underline{0.108}
& \textbf{0.123} & \underline{0.143} & \underline{0.127}
& \underline{74.70} & \underline{72.40} & \underline{86.82} \\
\ours GWM-g (Ours)
& \textbf{0.095} & \textbf{0.126} & \textbf{0.104}
& \textbf{0.123} & \textbf{0.139} & \textbf{0.125}
& \textbf{76.18} & \textbf{74.68} & \textbf{89.12} \\
\bottomrule
\end{tabular}
\end{table}

Table~\ref{tab:main-gaze-point} shows that the two observed-clip tasks separate the models differently.
On gaze localization, GWM matches the strongest image models: GWM-g is best or within 0.002 of the best on every GTE metric, and GWM-L matches PaGE on VideoAttentionTarget and is slightly behind it on ChildPlay.
On shift detection, in contrast, every GWM model outperforms every external baseline, and even GWM-B, with less than a tenth of the parameters of PaGE-H+, does so on all four shift metrics.
Accurate frame-wise gaze is therefore not sufficient to detect shifts.
Table~\ref{tab:ablation-prediction} locates the gain: about half of the 9.8-point lead in four-class AP over PaGE-H+ remains without the temporal trunk, and since the same frozen encoder without gaze supervision falls below both PaGE models, this half comes from person-conditioned gaze training rather than from the video encoder. The trunk and the world branch contribute the rest in similar parts. \looseness=-1

Frame-wise metrics cannot tell whether a model is stable because it tracks the gaze or because it suppresses movement. Table~\ref{tab:ablation-prediction-temporal} separates the two (Appendix~\ref{app:temporal-metrics}).
On both datasets, GWM-L and GWM-g jitter less during fixations than every external baseline and still follow the gaze path more closely when the gaze moves, so their stability does not come from smoothing.
Removing the temporal trunk loses both properties, which ties them to temporal modeling. The aligned residual measures path accuracy, not response latency. \looseness=-1

Table~\ref{tab:main-forecast} extends the comparison beyond the observed clip. GWM-L and GWM-g predict future gaze more accurately than all external baselines at both horizons, and forecast shifts with 8 to 10 points higher AP than the strongest baseline, SAGE with PaGE.
A strong current-gaze estimate alone could explain part of the localization gain, so we also compare with persistence, which repeats GWM-L's last observed estimate: forecasting lowers its future error by 15.6\% on VideoAttentionTarget and 11\% on ChildPlay. The model therefore anticipates where the gaze will move, not only where it is. \looseness=-1

\subsection{Ablation studies}
\label{sec:ablations}
Unless otherwise stated, we use GWM with a frozen V-JEPA~2.1 ViT-L encoder for all ablation experiments on VideoAttentionTarget and ChildPlay. Additional results are provided in Appendix~\ref{app:supp-experiments}.

\begin{table}[t]
\centering
\caption{Ablation of temporal modeling for gaze detection and forecasting. Future L2 uses the same frame-wise average as Table~\ref{tab:main-forecast}. Without the trunk, the world predictor has no input, so forecasting is not available (--). $-$world predictor removes the world branch and $\mathcal{L}_{\mathrm{lat}}$; future gaze is then read from the trunk. Last-window detection uses the observed-window shift probability as the forecasting score.}
\label{tab:ablation-prediction}

\footnotesize
\setlength{\tabcolsep}{5pt}
\renewcommand{\arraystretch}{1.05}
\definecolor{gazereference}{gray}{0.90}

\begin{tabular}{@{}l*{9}{c}@{}}
\toprule
& \multicolumn{4}{c}{\textbf{Gaze-target estimation}}
& \multicolumn{2}{c}{\footnotesize\textbf{Gaze-shift detection}}
& \multicolumn{3}{c}{\footnotesize\textbf{Future gaze shift}} \\
\cmidrule(lr){2-5}\cmidrule(lr){6-7}\cmidrule(lr){8-10}

\textbf{Variant}
& \multicolumn{2}{c}{\footnotesize\textbf{VAT}}
& \multicolumn{2}{c}{\footnotesize\textbf{ChildPlay}}
& \multicolumn{2}{c}{\footnotesize\textbf{Four-class}}
& \multicolumn{3}{c}{\footnotesize\textbf{Binary}} \\
\cmidrule(lr){2-3}\cmidrule(lr){4-5}\cmidrule(lr){6-7}\cmidrule(lr){8-10}

& {\footnotesize\shortstack{L2$\downarrow$\\Observed}}
& {\footnotesize\shortstack{L2$\downarrow$\\Future}}
& {\footnotesize\shortstack{L2$\downarrow$\\Observed}}
& {\footnotesize\shortstack{L2$\downarrow$\\Future}}
& {\footnotesize AP$\uparrow$}
& {\footnotesize F1$\uparrow$}
& {\footnotesize AP$\uparrow$}
& {\footnotesize F1$\uparrow$}
& {\footnotesize Recall$\uparrow$} \\
\midrule

\addlinespace[4pt]
$-$temporal trunk
& 0.073 & --
& 0.090 & --
& 71.25 & 69.97 & -- & -- & -- \\

$-$world predictor
& 0.072 & 0.124
& 0.087 & 0.135
& 73.92 & 73.03 & 71.69 & 70.64 & 84.18 \\

Last-window detection
& -- & --
& -- & --
& -- & -- & 63.47 & 67.83 & 75.39 \\

\rowcolor{gazereference}
$\lambda_{\mathrm{lat}}=0.1$
& \textbf{0.061} & \textbf{0.108}
& \textbf{0.073} & \underline{0.127}
& \textbf{76.09} & \textbf{75.05} & \textbf{74.70} & \textbf{72.40} & \textbf{86.82} \\

$\lambda_{\mathrm{lat}}=0.3$
& 0.067 & 0.113 
& \underline{0.078} & \textbf{0.125}
& 74.34 & 73.65 & 72.96 & 71.48 & \underline{85.29} \\

$\lambda_{\mathrm{lat}}=1.0$
& \underline{0.064} & \underline{0.110} 
& 0.081 & 0.129 
& \underline{74.82} & \underline{73.71} & \underline{73.40} & \underline{71.79} & 85.03 \\

\bottomrule
\end{tabular}
\end{table}

\noindent \textbf{Temporal modeling for detection and forecasting.}
Table~\ref{tab:ablation-prediction} evaluates the temporal trunk and latent world prediction
for observed gaze estimation, shift detection, and forecasting.
Removing either component worsens observed gaze-point accuracy on both datasets and lowers four-class AP/F1. Increasing the latent loss weight also weakens these results relative to the reference,
showing that stronger latent supervision does not necessarily improve gaze readout.
The largest tested weight still gives higher shift scores than removing the trunk,
so the temporal contribution is reduced rather than uniformly lost.
The reference weight, $\lambda_{\mathrm{lat}}=0.1$, gives the lowest observed gaze-point error and
highest shift scores among the tested settings, was selected on the validation splits, and is used in the remaining experiments.
The full observed-gaze metrics are reported in Appendix~\ref{app:full-temporal-ablation}. \looseness=-1

% Tables 5 and 6 share one float to save vertical space.
\begin{table}[t]
\footnotesize
\setlength{\tabcolsep}{3.5pt}
\renewcommand{\arraystretch}{1.05}
\definecolor{lossreference}{gray}{0.90}
\definecolor{gazereference}{gray}{0.90}
\begin{minipage}[t]{0.46\textwidth}
\centering
\caption{Ablation of the loss terms (shading: reference; --: no in/out head). Rows without $\mathcal{L}_{\mathrm{lat}}$ keep the world branch ($\lambda_{\mathrm{lat}}=0$); seeds in Appendix~\ref{app:seed-variability}.}
\label{tab:ablation-loss}
\fitwidth{%
\begin{tabular}{@{}*{11}{c}@{}}
\toprule
& & & \multicolumn{6}{c}{\textbf{Gaze-target estimation}}
& \multicolumn{2}{c}{\textbf{Gaze-shift det.}} \\
\cmidrule(lr){4-9}\cmidrule(lr){10-11}
\multicolumn{3}{c}{\textbf{Loss terms}}
& \multicolumn{3}{c}{\footnotesize\textbf{VideoAttentionTarget}}
& \multicolumn{3}{c}{\footnotesize\textbf{ChildPlay}}
& \multicolumn{2}{c}{\footnotesize\textbf{Four-class}} \\
\cmidrule(lr){1-3}\cmidrule(lr){4-6}\cmidrule(lr){7-9}\cmidrule(lr){10-11}

$\mathcal{L}_{\mathrm{heat}}$
& $\mathcal{L}_{\mathrm{inout}}$
& $\mathcal{L}_{\mathrm{lat}}$
& {\footnotesize AUC$\uparrow$}
& {\footnotesize L2$\downarrow$}
& {\footnotesize AP$\uparrow$}
& {\footnotesize AUC$\uparrow$}
& {\footnotesize L2$\downarrow$}
& {\footnotesize AP$\uparrow$}
& {\footnotesize AP$\uparrow$}
& {\footnotesize F1$\uparrow$} \\
\midrule

\rowcolor{lossreference}
\losstick & \losstick & \losstick
& \textbf{0.973} & \textbf{0.061} & \textbf{0.945}
& \textbf{0.968} & \textbf{0.073} & \underline{0.997}
& \textbf{76.09} & \textbf{75.05} \\

\addlinespace[3pt]
\losstick & \losstick & \losscross
& 0.963 & 0.072 & \underline{0.942}
& 0.960 & 0.087 & \textbf{0.999}
& 73.87 & 73.78 \\

\losstick & \losscross & \losscross
& 0.961 & 0.068 & --
& \underline{0.962} & 0.084 & --
& 72.71 & 71.90 \\

\losstick & \losscross & \losstick
& \underline{0.967} & \underline{0.065} & --
& \underline{0.962} & \underline{0.083} & --
& \underline{74.55} & \underline{74.38} \\
\bottomrule
\end{tabular}
}
\end{minipage}\hfill
\begin{minipage}[t]{0.50\textwidth}
\centering
\caption{Effect of unfreezing the encoder. Shading marks the frozen reference.}
\label{tab:ablation-unfreeze}
\fitwidth{%
\begin{tabular}{@{}l*{6}{c}@{}}
\toprule
& \multicolumn{4}{c}{\textbf{Gaze-target estimation}}
& \multicolumn{2}{c}{\textbf{Gaze-shift det.}} \\
\cmidrule(lr){2-5}\cmidrule(lr){6-7}

\textbf{Variant}
& \multicolumn{2}{c}{\textbf{VAT}}
& \multicolumn{2}{c}{\textbf{ChildPlay}}
& \multicolumn{2}{c}{\textbf{Four-class}} \\
\cmidrule(lr){2-3}\cmidrule(lr){4-5}\cmidrule(lr){6-7}

& AUC$\uparrow$
& L2$\downarrow$
& AUC$\uparrow$
& L2$\downarrow$
& AP$\uparrow$
& F1$\uparrow$ \\
\midrule

\shortstack[l]{Last 4 blocks\\unfrozen ($50.4$M)}
& \textbf{0.974} & 0.062
& 0.965 & 0.082
& 75.25 & 69.85 \\

\addlinespace[3pt]
\rowcolor{gazereference}
Frozen (reference)
& 0.973 & \textbf{0.061}
& \textbf{0.968} & \textbf{0.073}
& \textbf{76.09} & \textbf{75.05} \\
\bottomrule
\end{tabular}
}
\end{minipage}
\end{table}

\noindent \textbf{Loss terms of video training.}
Table~\ref{tab:ablation-loss} shows that combining in/out supervision and latent prediction gives the strongest shift detection. Compared with heatmap-only training, latent prediction improves both localization and four-class AP/F1; in/out supervision
alone improves shift scores but worsens localization. Combining both improves localization
on both datasets.
With capacity fixed, this comparison isolates the effect of the latent objective,
supporting its role in learning temporal representations for observed-clip detection.
Over three seeds, $\mathcal{L}_{\mathrm{lat}}$ improves four-class AP by 1.9 points ($p=0.04$), VAT L2 by 0.009 ($p=0.005$), and ChildPlay L2 by 0.010 ($p=0.004$), while its effect on four-class F1 remains within seed variation (Appendix~\ref{app:seed-variability}). \looseness=-1

\noindent \textbf{Unfreezing the video encoder.}
Table~\ref{tab:ablation-unfreeze} supports keeping the pretrained video encoder frozen
while adapting the temporal and task components. In the comparative run, the encoder
remains frozen for the first five epochs, after which its final four transformer blocks
and layer normalization are unfrozen and optimized with a reduced learning rate of
$1\times10^{-6}$ for the remaining fifteen epochs. Partial unfreezing provides no consistent
improvement in gaze estimation and weakens transition classification, with lower four-class
AP and macro-F1. We keep the encoder frozen throughout training, consistent with prior findings
in static gaze estimation \citep{ryan2025gazelle}. \looseness=-1

\begin{figure}[!h]
    \centering
    \includegraphics[width=0.8\textwidth]{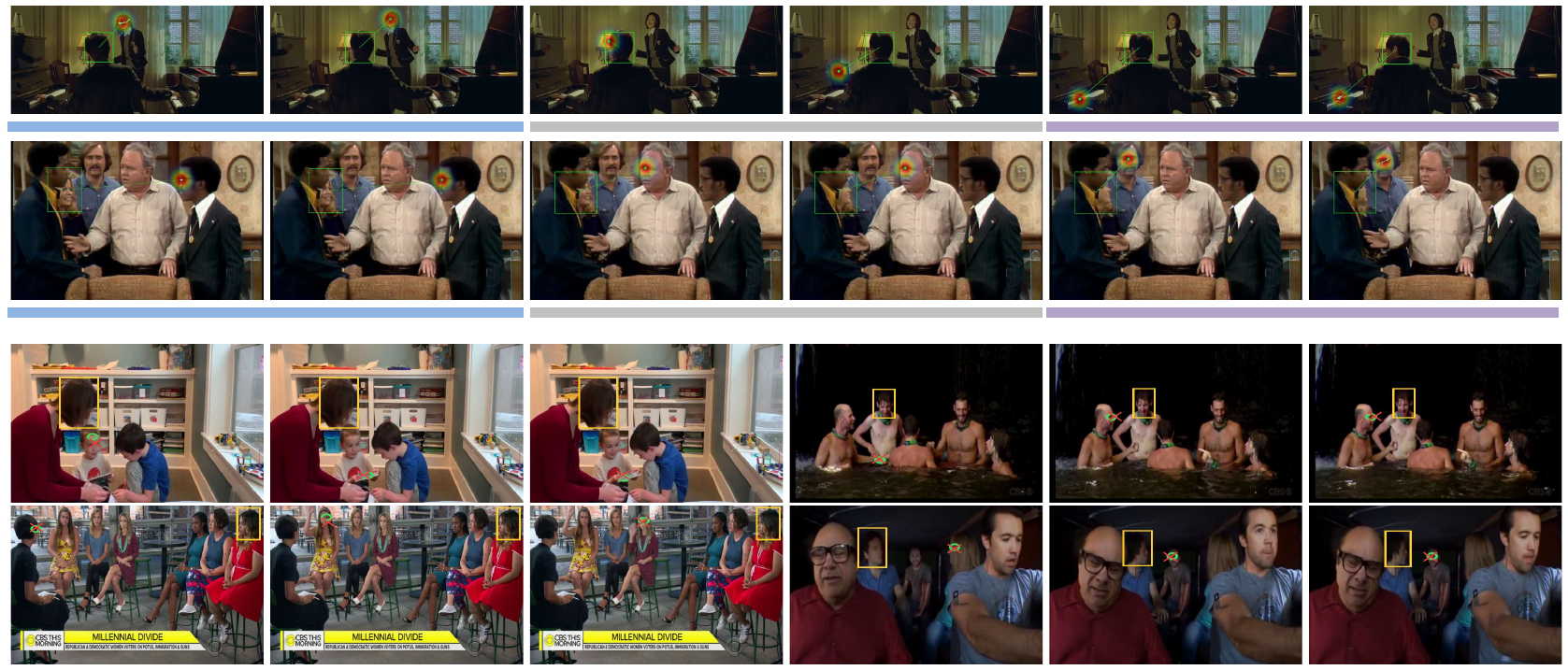}
    \caption{Gaze predictions of \modelname. Top: observed heatmaps; bars mark annotated fixation--transition--fixation intervals (blue--gray--purple). Bottom: four observed--future--future triplets. Yellow boxes mark subjects, green circles annotated gaze, and red crosses predictions. Future frames are shown for reference only.}
    \label{fig:qualitative_shifts}
\end{figure}

\noindent \textbf{Visualization.}
% \label{sec:visualization}
Figure~\ref{fig:qualitative_shifts} shows selected examples of observed and future gaze predictions. In the observed sequences, \modelname follows target changes in scenes with multiple people and partial occlusion. In the forecasting examples, predicted gaze points remain close to the annotated targets as attention shifts. More examples are in Appendix~\ref{app:qual}. \looseness=-1

% \subsection{Additional analyses}

% The supplementary material is organized to report the single-image and general-model comparison for Stage~2, full per-dataset metrics, temporal consistency, temporal-sampling and forecast-horizon sensitivity, performance by gaze-shift type, annotation quality, head-box robustness, computational cost, and qualitative failure cases as they become available. These analyses are separated from the main table so that the primary comparison remains focused on video GTE, video prediction, and gaze-shift forecasting.

% !TEX root = ../main.tex
\section{Conclusion}
\label{sec:conclusion}
We introduced the Gaze World Model (GWM), which models gaze shifts as, in part, responses to dynamic scene changes.
By co-training latent scene prediction with person-conditioned gaze supervision on a frozen world-model encoder, GWM links scene dynamics to individual attention.
This unified representation supports observed gaze estimation and shift detection, while a world predictor extends it to future forecasting.
We also augmented VideoAttentionTarget and ChildPlay with human-in-the-loop, event-level gaze-shift annotations.
Experiments show large gains in shift detection and forecasting alongside competitive gaze localization, suggesting that predicting how a scene evolves helps a model detect and anticipate gaze shifts.
\looseness=-1

\clearpage

\bibliography{references}

@inproceedings{recasens2015gazefollow,
  title={Where Are They Looking?},
  author={Recasens, Adri{\`a} and Khosla, Aditya and Vondrick, Carl and Torralba, Antonio},
  booktitle={Advances in Neural Information Processing Systems},
  volume={28},
  year={2015}
}

@inproceedings{chong2020vat,
  title={Detecting Attended Visual Targets in Video},
  author={Chong, Eunji and Wang, Yongxin and Ruiz, Nataniel and Rehg, James M.},
  booktitle={Proceedings of the IEEE/CVF Conference on Computer Vision and Pattern Recognition},
  pages={5395--5405},
  year={2020},
  doi={10.1109/CVPR42600.2020.00544}
}

@inproceedings{tafasca2023childplay,
  title={{ChildPlay}: A New Benchmark for Understanding Children's Gaze Behaviour},
  author={Tafasca, Samy and Gupta, Anshul and Odobez, Jean-Marc},
  booktitle={Proceedings of the IEEE/CVF International Conference on Computer Vision},
  pages={20878--20889},
  year={2023},
  doi={10.1109/ICCV51070.2023.01914}
}

@inproceedings{grauman2022ego4d,
  title={{Ego4D}: Around the World in 3,000 Hours of Egocentric Video},
  author={Grauman, Kristen and Westbury, Andrew and Byrne, Eugene and Chavis, Zachary and Furnari, Antonino and Girdhar, Rohit and Hamburger, Jackson and Jiang, Hao and Liu, Miao and Liu, Xingyu and Martin, Miguel and Nagarajan, Tushar and Radosavovic, Ilija and Ramakrishnan, Santhosh Kumar and Ryan, Fiona and Sharma, Jayant and Wray, Michael and Xu, Mengmeng and Xu, Eric Zhongcong and Zhao, Chen and Bansal, Siddhant and Batra, Dhruv and Cartillier, Vincent and Crane, Sean and Do, Tien and Doulaty, Morrie and Erapalli, Akshay and Feichtenhofer, Christoph and Fragomeni, Adriano and Fu, Qichen and Gebreselasie, Abrham and Gonz\'alez, Cristina and Hillis, James and Huang, Xuhua and Huang, Yifei and Jia, Wenqi and Khoo, Weslie and Kol\'a\v{r}, J\'achym and Kottur, Satwik and Kumar, Anurag and Landini, Federico and Li, Chao and Li, Yanghao and Li, Zhenqiang and Mangalam, Karttikeya and Modhugu, Raghava and Munro, Jonathan and Murrell, Tullie and Nishiyasu, Takumi and Price, Will and Ruiz Puentes, Paola and Ramazanova, Merey and Sari, Leda and Somasundaram, Kiran and Southerland, Audrey and Sugano, Yusuke and Tao, Ruijie and Vo, Minh and Wang, Yuchen and Wu, Xindi and Yagi, Takuma and Zhao, Ziwei and Zhu, Yunyi and Arbel\'aez, Pablo and Crandall, David and Damen, Dima and Farinella, Giovanni Maria and Fuegen, Christian and Ghanem, Bernard and Ithapu, Vamsi Krishna and Jawahar, C. V. and Joo, Hanbyul and Kitani, Kris and Li, Haizhou and Newcombe, Richard and Oliva, Aude and Park, Hyun Soo and Rehg, James M. and Sato, Yoichi and Shi, Jianbo and Shou, Mike Zheng and Torralba, Antonio and Torresani, Lorenzo and Yan, Mingfei and Malik, Jitendra},
  booktitle={Proceedings of the IEEE/CVF Conference on Computer Vision and Pattern Recognition},
  pages={18995--19012},
  year={2022}
}

@inproceedings{tafasca2024sharingan,
  title={Sharingan: A Transformer Architecture for Multi-Person Gaze Following},
  author={Tafasca, Samy and Gupta, Anshul and Odobez, Jean-Marc},
  booktitle={Proceedings of the IEEE/CVF Conference on Computer Vision and Pattern Recognition},
  pages={2008--2017},
  year={2024},
  doi={10.1109/CVPR52733.2024.00196}
}

@inproceedings{gupta2024mtgs,
  title={{MTGS}: A Novel Framework for Multi-Person Temporal Gaze Following and Social Gaze Prediction},
  author={Gupta, Anshul and Tafasca, Samy and Farkhondeh, Arya and Vuillecard, Pierre and Odobez, Jean-Marc},
  booktitle={Advances in Neural Information Processing Systems},
  volume={37},
  pages={15646--15673},
  year={2024},
  doi={10.52202/079017-0500}
}

@inproceedings{ryan2025gazelle,
  title={{Gaze-LLE}: Gaze Target Estimation via Large-Scale Learned Encoders},
  author={Ryan, Fiona and Bati, Ajay and Lee, Sangmin and Bolya, Daniel and Hoffman, Judy and Rehg, James M.},
  booktitle={Proceedings of the IEEE/CVF Conference on Computer Vision and Pattern Recognition},
  pages={28874--28884},
  year={2025},
  doi={10.1109/CVPR52734.2025.02689}
}

@inproceedings{miao2025multiview,
  title={Multi-View Gaze Target Estimation},
  author={Miao, Qiaomu and Golani, Vivek Raju and Xu, Jingyi and Dutta, Progga Paromita and Hoai, Minh and Samaras, Dimitris},
  booktitle={Proceedings of the IEEE/CVF International Conference on Computer Vision},
  pages={5371--5381},
  year={2025},
  doi={10.1109/ICCV51701.2025.00510}
}

@inproceedings{cao2026gazeanywhere,
  title={Gaze Target Estimation Anywhere with Concepts},
  author={Cao, Xu and Yang, Houze and Gunda, Vipin and Zhou, Zhongyi and Xu, Tianyu and Kowdle, Adarsh and Kim, Inki and Rehg, James M.},
  booktitle={Proceedings of the IEEE/CVF Conference on Computer Vision and Pattern Recognition},
  pages={31304--31315},
  year={2026}
}

@inproceedings{she2026textgaze,
  title={{TextGaze}: Prompting Gaze Target Estimation with Textual Scene Cues},
  author={She, Junhui and Wang, Fei and Li, Kun and Nie, Yiqi and Liu, Yuxin and Duan, Zhangling and Yang, Xun},
  booktitle={Proceedings of the Thirty-Fifth International Joint Conference on Artificial Intelligence},
  pages={1586--1594},
  year={2026},
  doi={10.24963/ijcai.2026/177}
}

@article{ye2026page,
  title={{PaGE}: Towards Practical Human-Level Gaze Target Estimation},
  author={Ye, Zhoutong and Zhang, Chengwen and Cui, Zhaibin and Sun, Mingze and Liu, Jiaqi and Li, Xiangwu and Wan, Qingyang and Liu, Chang and Wang, Xutong and Gao, Huan-ang and Mei, Yu and Yu, Chun and Shi, Yuanchun},
  journal={arXiv preprint arXiv:2607.04860},
  year={2026}
}

@inproceedings{miao2026omnigf,
  title={{OmniGF}: A Dual-Branch Vision-Language Framework for Unified Gaze Following},
  author={Miao, Qiaomu and Wu, Haoyu and Xu, Jingyi and Hoai, Minh and Samaras, Dimitris},
  booktitle={Advances in Neural Information Processing Systems},
  year={2026}
}

@article{li2026argaze,
  title={{ARGaze}: Autoregressive Transformers for Online Egocentric Gaze Estimation},
  author={Li, Jia and Zhao, Wenjie and Deng, Shijian and Lai, Bolin and Wu, Yuheng and Chen, Ruijia and Froehlich, Jon E. and Zhao, Yuhang and Tian, Yapeng},
  journal={arXiv preprint arXiv:2602.05132},
  year={2026}
}

@article{hessels2024fundamentals1,
  title={The Fundamentals of Eye Tracking Part 1: The Link Between Theory and Research Question},
  author={Hessels, Roy S. and Nuthmann, Antje and Nystr{\"o}m, Marcus and Andersson, Richard and Niehorster, Diederick C. and Hooge, Ignace T. C.},
  journal={Behavior Research Methods},
  volume={57},
  number={1},
  pages={16},
  year={2025},
  doi={10.3758/s13428-024-02544-8}
}

@article{hooge2025fundamentals2,
  title={The Fundamentals of Eye Tracking Part 2: From Research Question to Operationalization},
  author={Hooge, Ignace T. C. and Nuthmann, Antje and Nystr{\"o}m, Marcus and Niehorster, Diederick C. and Holleman, Gijs A. and Andersson, Richard and Hessels, Roy S.},
  journal={Behavior Research Methods},
  volume={57},
  number={2},
  pages={73},
  year={2025},
  doi={10.3758/s13428-024-02590-2}
}

@article{hooge2026fundamentals6,
  title={The Fundamentals of Eye Tracking Part 6: Working with Areas of Interest},
  author={Hooge, Ignace T. C. and Nystr{\"o}m, Marcus and Niehorster, Diederick C. and Andersson, Richard and Foulsham, Tom and Nuthmann, Antje and Hessels, Roy S.},
  journal={Behavior Research Methods},
  volume={58},
  number={3},
  pages={65},
  year={2026},
  doi={10.3758/s13428-025-02937-3}
}

@article{niehorster2026fundamentals7,
  title={The Fundamentals of Eye Tracking, Part 7: Determining Data Quality},
  author={Niehorster, Diederick C. and Nystr{\"o}m, Marcus and Hessels, Roy S. and Benjamins, Jeroen S. and Andersson, Richard and Hooge, Ignace T. C.},
  journal={Behavior Research Methods},
  volume={58},
  number={7},
  pages={183},
  year={2026},
  doi={10.3758/s13428-026-03039-4}
}

@inproceedings{assran2023ijepa,
  title={Self-Supervised Learning from Images with a Joint-Embedding Predictive Architecture},
  author={Assran, Mahmoud and Duval, Quentin and Misra, Ishan and Bojanowski, Piotr and Vincent, Pascal and Rabbat, Michael and LeCun, Yann and Ballas, Nicolas},
  booktitle={Proceedings of the IEEE/CVF Conference on Computer Vision and Pattern Recognition},
  pages={15619--15629},
  year={2023}
}

@article{assran2025vjepa2,
  title={{V-JEPA} 2: Self-Supervised Video Models Enable Understanding, Prediction and Planning},
  author={Assran, Mido and Bardes, Adrien and Fan, David and Garrido, Quentin and Howes, Russell and Komeili, Mojtaba and Muckley, Matthew and Rizvi, Ammar and Roberts, Claire and Sinha, Koustuv and Zholus, Artem and Arnaud, Sergio and Gejji, Abha and Martin, Ada and Hogan, Francois Robert and Dugas, Daniel and Bojanowski, Piotr and Khalidov, Vasil and Labatut, Patrick and Massa, Francisco and Szafraniec, Marc and Krishnakumar, Kapil and Li, Yong and Ma, Xiaodong and Chandar, Sarath and Meier, Franziska and LeCun, Yann and Rabbat, Michael and Ballas, Nicolas},
  journal={arXiv preprint arXiv:2506.09985},
  year={2025}
}

@inproceedings{murlabadia2026vjepa21,
  title={{V-JEPA} 2.1: Unlocking Dense Features in Video Self-Supervised Learning},
  author={Mur-Labadia, Lorenzo and Muckley, Matthew and Bar, Amir and Assran, Mido and Sinha, Koustuv and Rabbat, Mike and LeCun, Yann and Ballas, Nicolas and Bardes, Adrien},
  booktitle={European Conference on Computer Vision},
  pages={671--689},
  year={2026},
  publisher={Springer Nature Switzerland},
  doi={10.1007/978-3-032-37369-4_36}
}

@article{simeoni2025dinov3,
  title={{DINOv3}},
  author={Sim{\'e}oni, Oriane and others},
  journal={arXiv preprint arXiv:2508.10104},
  year={2025}
}

@inproceedings{carion2025sam3,
  title={{SAM 3}: Segment Anything with Concepts},
  author={Carion, Nicolas and Gustafson, Laura and Hu, Yuan-Ting and Debnath, Shoubhik and Hu, Ronghang and Suris Coll-Vinent, Didac and Ryali, Chaitanya and Alwala, Kalyan Vasudev and Khedr, Haitham and Huang, Andrew and Lei, Jie and Ma, Tengyu and Guo, Baishan and Kalla, Arpit and Marks, Markus and Greer, Joseph and Wang, Meng and Sun, Peize and R{\"a}dle, Roman and Afouras, Triantafyllos and Mavroudi, Effrosyni and Xu, Katherine and Wu, Tsung-Han and Zhou, Yu and Momeni, Liliane and Hazra, Rishi and Ding, Shuangrui and Vaze, Sagar and Porcher, Francois and Li, Feng and Li, Siyuan and Kamath, Aishwarya and Cheng, Ho Kei and Doll{\'a}r, Piotr and Ravi, Nikhila and Saenko, Kate and Zhang, Pengchuan and Feichtenhofer, Christoph},
  booktitle={International Conference on Learning Representations},
  year={2026}
}

@inproceedings{lian2018gaze,
  title={Believe It or Not, We Know What You Are Looking at!},
  author={Lian, Dongze and Yu, Zehao and Gao, Shenghua},
  booktitle={Proceedings of the Asian Conference on Computer Vision (ACCV)},
  pages={35--50},
  year={2018}
}

@inproceedings{chong2018connecting,
  title={Connecting Gaze, Scene, and Attention: Generalized Attention Estimation via Joint Modeling of Gaze and Scene Saliency},
  author={Chong, Eunji and Ruiz, Nataniel and Wang, Yongxin and Zhang, Yun and Rozga, Agata and Rehg, James M.},
  booktitle={Proceedings of the European Conference on Computer Vision (ECCV)},
  pages={397--412},
  year={2018}
}

@inproceedings{salvucci2000identifying,
  title={Identifying Fixations and Saccades in Eye-Tracking Protocols},
  author={Salvucci, Dario D. and Goldberg, Joseph H.},
  booktitle={Proceedings of the 2000 Symposium on Eye Tracking Research \& Applications (ETRA)},
  pages={71--78},
  year={2000}
}

@book{holmqvist2011eye,
  title={Eye Tracking: A Comprehensive Guide to Methods and Measures},
  author={Holmqvist, Kenneth and Nystr{\"o}m, Marcus and Andersson, Richard and Dewhurst, Richard and Jarodzka, Halszka and Van de Weijer, Joost},
  year={2011},
  publisher={Oxford University Press},
  address={Oxford, UK},
  isbn={9780199697083}
}

@article{blignaut2009fixation,
  title={Fixation Identification: The Optimum Threshold for a Dispersion Algorithm},
  author={Blignaut, Pieter},
  journal={Attention, Perception, \& Psychophysics},
  volume={71},
  number={4},
  pages={881--895},
  year={2009},
  publisher={Springer},
  doi={10.3758/APP.71.4.881}
}

@article{bardes2024vjepa,
  title={Revisiting Feature Prediction for Learning Visual Representations from Video},
  author={Bardes, Adrien and Garrido, Quentin and Ponce, Jean and Chen, Xinlei and Rabbat, Michael and LeCun, Yann and Assran, Mahmoud and Ballas, Nicolas},
  journal={Transactions on Machine Learning Research},
  year={2024}
}

@article{hayhoe2005eye,
  title={Eye Movements in Natural Behavior},
  author={Hayhoe, Mary and Ballard, Dana},
  journal={Trends in Cognitive Sciences},
  volume={9},
  number={4},
  pages={188--194},
  year={2005},
  publisher={Elsevier},
  doi={10.1016/j.tics.2005.02.009}
}

@article{land2006eyemovements,
  title={Eye Movements and the Control of Actions in Everyday Life},
  author={Land, Michael F.},
  journal={Progress in Retinal and Eye Research},
  volume={25},
  number={3},
  pages={296--324},
  year={2006},
  doi={10.1016/j.preteyeres.2006.01.002}
}

@article{huang2015gazeintent,
  title={Using Gaze Patterns to Predict Task Intent in Collaboration},
  author={Huang, Chien-Ming and Andrist, Sean and Saupp{\'e}, Allison and Mutlu, Bilge},
  journal={Frontiers in Psychology},
  volume={6},
  pages={1049},
  year={2015},
  doi={10.3389/fpsyg.2015.01049}
}

@article{frischen2007gazecueing,
  title={Gaze Cueing of Attention: Visual Attention, Social Cognition, and Individual Differences},
  author={Frischen, Alexandra and Bayliss, Andrew P. and Tipper, Steven P.},
  journal={Psychological Bulletin},
  volume={133},
  number={4},
  pages={694--724},
  year={2007},
  doi={10.1037/0033-2909.133.4.694}
}

@inproceedings{kuang2026sage,
  title={{SAGE}: Synchronized Action-Gaze Recognition and Anticipation for Human Behavior Understanding},
  author={Kuang, Chenyi and Agarwal, Nakul},
  booktitle={Proceedings of the European Conference on Computer Vision},
  year={2026}
}

@inproceedings{zhang2017dfg,
  title={Deep Future Gaze: Gaze Anticipation on Egocentric Videos Using Adversarial Networks},
  author={Zhang, Mengmi and Ma, Keng Teck and Lim, Joo Hwee and Zhao, Qi and Feng, Jiashi},
  booktitle={Proceedings of the IEEE Conference on Computer Vision and Pattern Recognition},
  pages={4372--4381},
  year={2017}
}

@inproceedings{li2018eyebeholder,
  title={In the Eye of Beholder: Joint Learning of Gaze and Actions in First Person Video},
  author={Li, Yin and Liu, Miao and Rehg, James M.},
  booktitle={Proceedings of the European Conference on Computer Vision},
  pages={639--655},
  year={2018},
  doi={10.1007/978-3-030-01228-1_38}
}

@inproceedings{lai2022glc,
  title={In the Eye of Transformer: Global-Local Correlation for Egocentric Gaze Estimation},
  author={Lai, Bolin and Liu, Miao and Ryan, Fiona and Rehg, James M.},
  booktitle={British Machine Vision Conference},
  year={2022}
}

@article{furnari2020rulstm,
  title={Rolling-Unrolling {LSTM}s for Action Anticipation from First-Person Video},
  author={Furnari, Antonino and Farinella, Giovanni Maria},
  journal={IEEE Transactions on Pattern Analysis and Machine Intelligence},
  volume={43},
  number={11},
  pages={4021--4036},
  year={2021},
  doi={10.1109/TPAMI.2020.2992889}
}

@article{ha2018worldmodels,
  title={World Models},
  author={Ha, David and Schmidhuber, J{\"u}rgen},
  journal={arXiv preprint arXiv:1803.10122},
  year={2018}
}

@article{hafner2023dreamerv3,
  title={Mastering Diverse Control Tasks through World Models},
  author={Hafner, Danijar and Pasukonis, Jurgis and Ba, Jimmy and Lillicrap, Timothy},
  journal={Nature},
  volume={640},
  number={8059},
  pages={647--653},
  year={2025},
  doi={10.1038/s41586-025-08744-2}
}

@article{shen2026wam,
  title={World Action Models: A Survey},
  author={Shen, Qiuhong and Zhang, Shihua and Liao, Yue and Li, Qi and Tan, Zhenxiong and Wang, Shizun and Yan, Shuicheng and Wang, Xinchao},
  journal={arXiv preprint arXiv:2606.20781},
  year={2026}
}

@inproceedings{vondrick2016anticipating,
  title={Anticipating Visual Representations from Unlabeled Video},
  author={Vondrick, Carl and Pirsiavash, Hamed and Torralba, Antonio},
  booktitle={Proceedings of the IEEE Conference on Computer Vision and Pattern Recognition},
  pages={98--106},
  year={2016}
}

@inproceedings{girdhar2021avt,
  title={Anticipative Video Transformer},
  author={Girdhar, Rohit and Grauman, Kristen},
  booktitle={Proceedings of the IEEE/CVF International Conference on Computer Vision},
  pages={13485--13495},
  year={2021}
}

@inproceedings{tonini2023objectaware,
  title={Object-aware Gaze Target Detection},
  author={Tonini, Francesco and Dall'Asen, Nicola and Beyan, Cigdem and Ricci, Elisa},
  booktitle={Proceedings of the IEEE/CVF International Conference on Computer Vision},
  pages={21860--21869},
  year={2023}
}

@inproceedings{wang2022gatector,
  title={{GaTector}: A Unified Framework for Gaze Object Prediction},
  author={Wang, Binglu and Hu, Tao and Li, Baoshan and Chen, Xiaojuan and Zhang, Zhijie},
  booktitle={Proceedings of the IEEE/CVF Conference on Computer Vision and Pattern Recognition},
  pages={19566--19575},
  year={2022}
}

@article{ding2024worldmodelsurvey,
  title={Understanding World or Predicting Future? A Comprehensive Survey of World Models},
  author={Ding, Jingtao and Zhang, Yunke and Shang, Yu and Zhang, Yuheng and Zong, Zefang and Feng, Jie and Yuan, Yuan and Su, Hongyuan and Li, Nian and Sukiennik, Nicholas and Xu, Fengli and Li, Yong},
  journal={ACM Computing Surveys},
  volume={58},
  number={3},
  pages={1--38},
  year={2025},
  doi={10.1145/3746449}
}

@inproceedings{huang2018attentiontransition,
  title={Predicting Gaze in Egocentric Video by Learning Task-dependent Attention Transition},
  author={Huang, Yifei and Cai, Minjie and Li, Zhenqiang and Sato, Yoichi},
  booktitle={Proceedings of the European Conference on Computer Vision},
  pages={789--804},
  year={2018}
}

@inproceedings{mendonca2023structured,
  title={Structured World Models from Human Videos},
  author={Mendonca, Russell and Bahl, Shikhar and Pathak, Deepak},
  booktitle={Robotics: Science and Systems},
  year={2023}
}

@inproceedings{chen2025egoagent,
  title={{EgoAgent}: A Joint Predictive Agent Model in Egocentric Worlds},
  author={Chen, Lu and Wang, Yizhou and Tang, Shixiang and Ma, Qianhong and He, Tong and Ouyang, Wanli and Zhou, Xiaowei and Bao, Hujun and Peng, Sida},
  booktitle={Proceedings of the IEEE/CVF International Conference on Computer Vision},
  pages={6970--6980},
  year={2025}
}

@misc{qwenteam2026qwen38,
  title={{Qwen3.8-27B}},
  author={{Qwen Team}},
  howpublished={Online},
  year={2026},
  url={https://huggingface.co/Qwen/Qwen3.8-27B}
}

@misc{geminiteam2026gemini38,
  title={{Gemini 3.8 Flash} Model Card},
  author={{Google DeepMind}},
  howpublished={Online},
  year={2026},
  url={https://deepmind.google/models/model-cards/gemini-3-8-flash/}
}

@inproceedings{wang2025actiondetail,
  title={Action Detail Matters: Refining Video Recognition with Local Action Queries},
  author={Wang, Mengmeng and Huang, Zeyi and Kong, Xiangjie and Shen, Guojiang and Dai, Guang and Wang, Jingdong and Liu, Yong},
  booktitle={Proceedings of the IEEE/CVF Conference on Computer Vision and Pattern Recognition},
  pages={19132--19142},
  year={2025}
}

@article{tomasello1986joint,
  title={Joint attention and early language},
  author={Tomasello, Michael and Farrar, Michael Jeffrey},
  journal={Child Development},
  volume={57},
  number={6},
  pages={1454--1463},
  year={1986},
  doi={10.2307/1130423}
}

@inproceedings{recasens2017following,
  title={Following Gaze in Video},
  author={Recasens, Adri{\`a} and Vondrick, Carl and Khosla, Aditya and Torralba, Antonio},
  booktitle={Proceedings of the IEEE International Conference on Computer Vision},
  pages={1444--1452},
  year={2017},
  doi={10.1109/ICCV.2017.160}
}

@inproceedings{miao2023patch,
  title={Patch-Level Gaze Distribution Prediction for Gaze Following},
  author={Miao, Qiaomu and Hoai, Minh and Samaras, Dimitris},
  booktitle={Proceedings of the IEEE/CVF Winter Conference on Applications of Computer Vision},
  pages={880--889},
  year={2023},
  doi={10.1109/WACV56688.2023.00094}
}

@inproceedings{fang2021dual,
  title={Dual Attention Guided Gaze Target Detection in the Wild},
  author={Fang, Yi and Tang, Jiapeng and Shen, Wang and Shen, Wei and Gu, Xiao and Song, Li and Zhai, Guangtao},
  booktitle={Proceedings of the IEEE/CVF Conference on Computer Vision and Pattern Recognition},
  pages={11390--11399},
  year={2021}
}

@inproceedings{bao2022escnet,
  title={{ESCNet}: Gaze Target Detection with the Understanding of {3D} Scenes},
  author={Bao, Jun and Liu, Buyu and Yu, Jun},
  booktitle={Proceedings of the IEEE/CVF Conference on Computer Vision and Pattern Recognition},
  pages={14106--14115},
  year={2022}
}

@inproceedings{tu2022endtoend,
  title={End-to-End Human-Gaze-Target Detection with Transformers},
  author={Tu, Danyang and Min, Xiongkuo and Duan, Huiyu and Guo, Guodong and Zhai, Guangtao and Shen, Wei},
  booktitle={Proceedings of the IEEE/CVF Conference on Computer Vision and Pattern Recognition},
  pages={2192--2200},
  year={2022}
}

@inproceedings{tafasca2024semantic,
  title={Toward Semantic Gaze Target Detection},
  author={Tafasca, Samy and Gupta, Anshul and Bros, Victor and Odobez, Jean-Marc},
  booktitle={Advances in Neural Information Processing Systems},
  volume={37},
  pages={121422--121448},
  year={2024},
  doi={10.52202/079017-3858}
}

@article{song2024vitgaze,
  title={{ViTGaze}: Gaze Following with Interaction Features in Vision Transformers},
  author={Song, Yuehao and Wang, Xinggang and Yao, Jingfeng and Liu, Wenyu and Zhang, Jinglin and Xu, Xiangmin},
  journal={Visual Intelligence},
  volume={2},
  number={1},
  pages={31},
  year={2024},
  doi={10.1007/s44267-024-00064-9}
}

@inproceedings{fan2018sharedattention,
  title={Inferring Shared Attention in Social Scene Videos},
  author={Fan, Lifeng and Chen, Yixin and Wei, Ping and Wang, Wenguan and Zhu, Song-Chun},
  booktitle={Proceedings of the IEEE Conference on Computer Vision and Pattern Recognition},
  pages={6460--6468},
  year={2018},
  doi={10.1109/CVPR.2018.00676}
}

@inproceedings{fan2019gazecommunication,
  title={Understanding Human Gaze Communication by Spatio-Temporal Graph Reasoning},
  author={Fan, Lifeng and Wang, Wenguan and Huang, Siyuan and Tang, Xinyu and Zhu, Song-Chun},
  booktitle={Proceedings of the IEEE/CVF International Conference on Computer Vision},
  pages={5723--5732},
  year={2019}
}

@inproceedings{marinjimenez2019laeonet,
  title={{LAEO-Net}: Revisiting People Looking at Each Other in Videos},
  author={Marin-Jimenez, Manuel J. and Kalogeiton, Vicky and Medina-Suarez, Pablo and Zisserman, Andrew},
  booktitle={Proceedings of the IEEE/CVF Conference on Computer Vision and Pattern Recognition},
  pages={3477--3485},
  year={2019}
}

@article{zemblys2019gazenet,
  title={{gazeNet}: End-to-End Eye-Movement Event Detection with Deep Neural Networks},
  author={Zemblys, Raimondas and Niehorster, Diederick C. and Holmqvist, Kenneth},
  journal={Behavior Research Methods},
  volume={51},
  number={2},
  pages={840--864},
  year={2019},
  doi={10.3758/s13428-018-1133-5}
}

@article{startsev2019cnnblstm,
  title={{1D CNN} with {BLSTM} for Automated Classification of Fixations, Saccades, and Smooth Pursuits},
  author={Startsev, Mikhail and Agtzidis, Ioannis and Dorr, Michael},
  journal={Behavior Research Methods},
  volume={51},
  number={2},
  pages={556--572},
  year={2019},
  doi={10.3758/s13428-018-1144-2}
}

@article{emery2000eyes,
  title={The Eyes Have It: The Neuroethology, Function and Evolution of Social Gaze},
  author={Emery, Nathan J.},
  journal={Neuroscience \& Biobehavioral Reviews},
  volume={24},
  number={6},
  pages={581--604},
  year={2000},
  doi={10.1016/S0149-7634(00)00025-7}
}

@article{scaife1975capacity,
  title={The Capacity for Joint Visual Attention in the Infant},
  author={Scaife, Michael and Bruner, Jerome S.},
  journal={Nature},
  volume={253},
  number={5489},
  pages={265--266},
  year={1975},
  doi={10.1038/253265a0}
}

@article{mundy1986defining,
  title={Defining the Social Deficits of Autism: The Contribution of Non-Verbal Communication Measures},
  author={Mundy, Peter and Sigman, Marian and Ungerer, Judy and Sherman, Tracy},
  journal={Journal of Child Psychology and Psychiatry},
  volume={27},
  number={5},
  pages={657--669},
  year={1986},
  doi={10.1111/j.1469-7610.1986.tb00190.x}
}

@article{chong2020eyecontact,
  title={Detection of Eye Contact with Deep Neural Networks Is as Accurate as Human Experts},
  author={Chong, Eunji and Clark-Whitney, Elysha and Southerland, Audrey and Stubbs, Elizabeth and Miller, Chanel and Ajodan, Eliana L. and Silverman, Melanie R. and Lord, Catherine and Rozga, Agata and Jones, Rebecca M. and Rehg, James M.},
  journal={Nature Communications},
  volume={11},
  number={1},
  pages={6386},
  year={2020},
  doi={10.1038/s41467-020-19712-x}
}

@article{jones2013attention,
  title={Attention to Eyes Is Present but in Decline in 2--6-Month-Old Infants Later Diagnosed with Autism},
  author={Jones, Warren and Klin, Ami},
  journal={Nature},
  volume={504},
  number={7480},
  pages={427--431},
  year={2013},
  doi={10.1038/nature12715}
}

@inproceedings{zhou2025dinowm,
  title={{DINO-WM}: World Models on Pre-trained Visual Features enable Zero-shot Planning},
  author={Zhou, Gaoyue and Pan, Hengkai and LeCun, Yann and Pinto, Lerrel},
  booktitle={Proceedings of the 42nd International Conference on Machine Learning},
  series={Proceedings of Machine Learning Research},
  volume={267},
  pages={79115--79135},
  year={2025}
}

@inproceedings{liang2019peeking,
  title={Peeking into the Future: Predicting Future Person Activities and Locations in Videos},
  author={Liang, Junwei and Jiang, Lu and Niebles, Juan Carlos and Hauptmann, Alexander G. and Fei-Fei, Li},
  booktitle={Proceedings of the IEEE/CVF Conference on Computer Vision and Pattern Recognition},
  pages={5718--5727},
  year={2019}
}

@inproceedings{barquero2022didnt,
  title={Didn't See That Coming: A Survey on Non-verbal Social Human Behavior Forecasting},
  author={Barquero, German and N{\'u}{\~n}ez, Johnny and Escalera, Sergio and Xu, Zhen and Tu, Wei-Wei and Guyon, Isabelle and Palmero, Cristina},
  booktitle={Understanding Social Behavior in Dyadic and Small Group Interactions},
  series={Proceedings of Machine Learning Research},
  volume={173},
  pages={139--178},
  year={2022}
}

@inproceedings{vaswani2017attention,
  title={Attention Is All You Need},
  author={Vaswani, Ashish and Shazeer, Noam and Parmar, Niki and Uszkoreit, Jakob and Jones, Llion and Gomez, Aidan N. and Kaiser, {\L}ukasz and Polosukhin, Illia},
  booktitle={Advances in Neural Information Processing Systems},
  volume={30},
  year={2017}
}

@inproceedings{dosovitskiy2021vit,
  title={An Image is Worth 16x16 Words: Transformers for Image Recognition at Scale},
  author={Dosovitskiy, Alexey and Beyer, Lucas and Kolesnikov, Alexander and Weissenborn, Dirk and Zhai, Xiaohua and Unterthiner, Thomas and Dehghani, Mostafa and Minderer, Matthias and Heigold, Georg and Gelly, Sylvain and Uszkoreit, Jakob and Houlsby, Neil},
  booktitle={International Conference on Learning Representations},
  year={2021}
}

@inproceedings{zhai2022scaling,
  title={Scaling Vision Transformers},
  author={Zhai, Xiaohua and Kolesnikov, Alexander and Houlsby, Neil and Beyer, Lucas},
  booktitle={Proceedings of the IEEE/CVF Conference on Computer Vision and Pattern Recognition},
  pages={12104--12113},
  year={2022}
}

@inproceedings{lin2017focal,
  title={Focal Loss for Dense Object Detection},
  author={Lin, Tsung-Yi and Goyal, Priya and Girshick, Ross and He, Kaiming and Doll{\'a}r, Piotr},
  booktitle={Proceedings of the IEEE International Conference on Computer Vision},
  pages={2980--2988},
  year={2017}
}

@misc{geminiteam2026gemini35,
  title={{Gemini 3.5 Flash} Model Card},
  author={{Google DeepMind}},
  howpublished={Online},
  year={2026},
  url={https://deepmind.google/models/model-cards/gemini-3-5-flash/}
}
\bibliographystyle{plainnat_iclr}

% !TEX root = ../main.tex
\appendix
% The appendix uses standard LaTeX float and caption spacing.
\setlength{\textfloatsep}{20pt plus 2pt minus 4pt}
\setlength{\floatsep}{12pt plus 2pt minus 2pt}
\setlength{\intextsep}{12pt plus 2pt minus 2pt}
\setlength{\dbltextfloatsep}{20pt plus 2pt minus 4pt}
\setlength{\dblfloatsep}{12pt plus 2pt minus 2pt}
\setlength{\abovecaptionskip}{10pt}
\clearpage

\section*{Appendix}\label{sec:appendix}
\crefalias{section}{appendix}

\startcontents[sections]
\printcontents[sections]{l}{1}{\setcounter{tocdepth}{2}}

\clearpage

% Appendix order follows the main text: related work, model details (Section 3), data (Section 4),
% evaluation metrics and supplementary experiments (Section 5), qualitative
% examples, then limitations.
% !TEX root = ../main.tex
\section{Extended Related Work}
\label{app:extended-related-work}

\subsection{Gaze-event definitions and eye-tracking methodology}
A gaze event is defined operationally rather than read directly from a raw signal. Established eye-tracking methodology separates continuous ocular kinematics from discrete behavioral events \citep{salvucci2000identifying,holmqvist2011eye,hessels2024fundamentals1}. Classical velocity- and dispersion-threshold methods isolate fixations and saccades from point-of-regard streams \citep{salvucci2000identifying,blignaut2009fixation}. Learned detectors replace hand-tuned thresholds, classifying eye-tracker samples into fixations, saccades, and post-saccadic oscillations end to end from raw data \citep{zemblys2019gazenet}, or into fixations, saccades, and smooth pursuit from raw coordinates and multi-scale velocity features \citep{startsev2019cnnblstm}. In dynamic scenes, however, target transitions also depend on moving object boundaries, scene geometry, and target visibility \citep{hayhoe2005eye,niehorster2026fundamentals7}, motivating criteria such as minimum dwell times, transition-duration bounds, and measurement noise floors \citep{hooge2025fundamentals2,hooge2026fundamentals6}. These protocols assume high-frequency eye-tracker signals. The third-person datasets used here do not provide synchronized high-frequency eye-tracking signals, so target-level events must instead be defined through the temporal persistence of target identity, supported by instance tracking and semantic verification. Out-of-frame gaze must likewise be treated as an explicit visibility state rather than missing data, which does not by itself distinguish one off-screen target from another \citep{chong2020vat,tafasca2023childplay}.

\subsection{Gaze and action anticipation}
Visual anticipation has generally been formulated as predicting a future gaze coordinate or an upcoming action label. In egocentric vision, early work forecast prospective fixation points from wearable-camera streams \citep{zhang2017dfg}; later methods connected gaze to action recognition and anticipation \citep{li2018eyebeholder,lai2022glc} or modeled gaze trajectories autoregressively \citep{li2026argaze}. SAGE jointly models human--object interactions and visual attention across egocentric and exocentric views \citep{kuang2026sage}, while large-scale video benchmarks define action anticipation through symbolic verb--noun categories \citep{grauman2022ego4d}, which anticipation models predict as class labels \citep{furnari2020rulstm,girdhar2021avt}. We complement gaze-coordinate prediction with explicit person-involving transition labels for a specified person. Detection classifies these transitions in observed video, while forecasting predicts whether one will occur in the future.

\subsection{Video gaze architectures and subject conditioning}
Gaze target models have largely extended frame-wise gaze estimation by aggregating independently encoded frame features. Recurrent temporal processing supports out-of-frame gaze prediction \citep{chong2020vat}, while later systems incorporate multi-agent reasoning \citep{tafasca2024sharingan}, temporal-social classification \citep{gupta2024mtgs}, multi-view fusion \citep{miao2025multiview}, and large pretrained encoders, frozen \citep{ryan2025gazelle} or fine-tuned \citep{ye2026page}. Subject specification has also expanded beyond head bounding boxes to textual and multimodal prompts \citep{cao2026gazeanywhere,she2026textgaze,miao2026omnigf}. Nevertheless, in our experiments, image models applied frame by frame produce more temporal jitter than \modelname{} (Table~\ref{tab:ablation-prediction-temporal}), and post-hoc smoothing can suppress genuine rapid transitions \citep{li2026argaze}. This motivates learning temporal representations before decoding gaze rather than treating temporal context solely as an aggregation or filtering step.

% !TEX root = ../main.tex
\section{Model Details}
\label{app:model-details}

\textbf{Head-crop input.}
For each observed frame, the head crop of the specified person is cut from the frame using its head box and resized to $256\times256$. The head-crop clip enters SIM as a separate head stream; with the same two-frame tubelets, each head tubelet yields $P_h=16\times16=256$ tokens. SIM projects these tokens to width $D$ and places them inside the head-box rectangle through their RoPE coordinates (Section~\ref{sec:architecture}).

\noindent\label{app:baseline-adaptation}\textbf{Baseline adaptation.}
Gaze-LLE and PaGE are image models without a shift-detection output. Gaze-LLE counts in Table~\ref{tab:main-gaze-point} include only the frozen encoder; with its 2.8M/2.9M decoder, the totals are 89M (ViT-B) and 307M (ViT-L). MTGS \citep{gupta2024mtgs} aggregates time within the gaze model and predicts per-frame gaze heatmaps, in/out labels, and pairwise social-gaze labels, including looking at humans (LAH), but no gaze-shift events; we use its released checkpoint on both datasets. We keep each model frozen and extract, per frame, its person-conditioned scene features (after SIM for PaGE; the input to the heatmap head for Gaze-LLE). These features are pooled with the model's own predicted heatmap and concatenated with its predicted gaze point, in/out probability, and the head box; no ground-truth gaze enters this input. A trainable temporal classifier (two Transformer layers with sinusoidal time encoding, masked mean pooling, and an MLP) maps the per-frame vectors to the four window classes.

For the rule engine on MTGS, the Stage-1 rules of Appendix~\ref{app:data-engine} are applied to the MTGS predictions with one change: in each frame where the highest LAH score of the specified person toward another person exceeds the decision threshold, the target is that person; the remaining in-frame targets are assigned from the MTGS gaze point and the SAM~3 masks as in Stage~1.

SAGE has no public code, so we re-implement its gaze-anticipation branch without the action branch, which requires action labels that VAT and ChildPlay lack. Its human--object context tokens are replaced by pooled scene tokens of the gaze detector, and one anticipation step corresponds to one two-frame tubelet. The gaze detector is either Sharingan, initialized from its released VAT checkpoint, or PaGE ViT-B with a DINOv3 backbone~\citep{simeoni2025dinov3}, and is applied frame by frame. For SAGE with Sharingan, the detector is frozen and only the anticipation module is trained. For SAGE with PaGE, the detector is fine-tuned jointly with the anticipation module following SAGE's two-step schedule: 5 epochs on current-gaze labels, then 20 epochs with future-gaze labels added. SAGE has no shift output, so for binary shift forecasting we freeze the trained SAGE model and train a shift head on its per-step anticipation tokens, with the same training data, loss, and optimization as the forecasting head of \modelname.

% !TEX root = ../main.tex
\section{Gaze-Shift Data Engine}
\label{app:data-engine}
\label{sec:pipeline-stages}

\begin{figure}[h]
    \centering
    \includegraphics[width=0.9\textwidth]{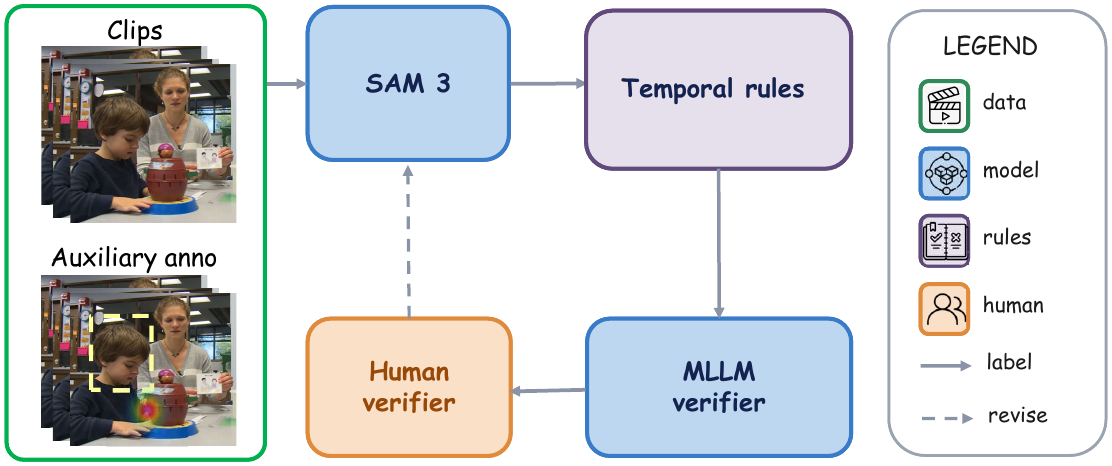}
    \caption{The three-stage gaze-shift data engine.}
    \label{fig:data-engine}
\end{figure}

Table~\ref{tab:gaze-task-terms} defines the terms used in Section~\ref{sec:GWM-task}.
We then detail the data engine summarized in Section~\ref{sec:pipeline} and
Figure~\ref{fig:data-engine}.

\begin{table}[t]
\centering
\caption{Terminology for annotated events, benchmark labels, and model outputs.}
\label{tab:gaze-task-terms}
\small
\setlength{\tabcolsep}{4pt}
\begin{tabular}{@{}p{0.24\linewidth}p{\dimexpr0.76\linewidth-8pt\relax}@{}}
\toprule
\textbf{Term} & \textbf{Meaning in this work} \\
\midrule
Semantic target & The person, object, or region to which gaze is assigned; object keys do not guarantee persistent instance identity. \\
Person identity & A person identifier propagated during annotation, distinguishing different people from body parts of the same person. \\
\texttt{other} & Objects, background regions, and out-of-frame gaze; it does not identify a particular off-screen target. \\
Held object & An item held, worn, or manipulated by a person. Its key records the identifier of that person, for example \texttt{p001:phone}, but the target counts as \texttt{other} in window labels. \\
Hard negative & A candidate that the semantic verifier rejects as a shift; it is retained as a negative example for training. \\
Event onset / offset & The annotated start and end of the transition interval between stable targets $A$ and $B$. \\
Window label & A gaze-shift event is assigned to a window if the transition starts within that window, even if it ends after the window. Observed windows have four labels; future windows have binary labels. \\
No in-scope shift & A valid window contains none of the three person-involving transition types. \\
Unverified & Event status or required annotations remain unresolved after review; affected windows are excluded, not labeled negative. \\
\bottomrule
\end{tabular}
\end{table}

\paragraph{Stage 1: rule-based candidate generation.}
For each annotated person and frame, the rule stage assigns a gaze target. Out-of-frame status is taken from the source annotation, the coordinate $(-1,-1)$ in VAT and the \texttt{outside\_frame} state in ChildPlay, and is never inferred from a missing coordinate. For in-frame gaze, SAM~3~\citep{carion2025sam3} provides human-instance masks from a ``person'' text prompt and a target mask from the annotated gaze point used as a point prompt. A target that overlaps a person mask becomes a human target with that person's identifier, and any other target becomes an object target. The resulting sequence of target keys is run-length encoded, and runs of the same target separated by a short gap are merged. Two stable runs that each last at least $T_{\text{dwell}}$ and are separated by a transition of at most $T_{\text{trans}}$ form a candidate shift (Table~\ref{tab:rule-thresholds}). The rule stage abstains when mask agreement is low, when more than half of the frames have low confidence, or when more than 30\% of the frames lack a mask.

\begin{table}[b]
\centering
\caption{Main rule-filter thresholds.}
\label{tab:rule-thresholds}
\small
\begin{tabular}{lll}
\toprule
\textbf{Parameter} & \textbf{Value} & \textbf{Function} \\
\midrule
Stable fixation ($T_{\text{dwell}}$) & 133 / 150\,ms & fixation per endpoint (VAT / ChildPlay) \\
Maximum transition ($T_{\text{trans}}$) & 1,000\,ms & maximum interval between stable endpoints \\
Same-target merge gap & 67\,ms & maximum gap for merging runs of one target \\
\bottomrule
\end{tabular}
\end{table}

\paragraph{Stage 2: multimodal verification.}
\label{app:semantic-verification}
Each candidate is rendered as a contact sheet whose tiles are grouped into five segments: frames before, during, and after the transition, and two context segments that show the target attended before the event and any later return to it. Each tile overlays the head box of the specified person, colored by the annotated in/out state, the annotated gaze point, the SAM~3 target contour, and the masks and identifiers of the other people. Gemini~3.5~Flash~\citep{geminiteam2026gemini35} receives the contact sheet together with a prompt that contains the Stage-1 fields of the candidate, namely the targets before, during, and after the transition, their person identifiers, the transition type, and the dwell times. It returns one JSON object constrained by a response schema; decoding uses temperature 0, a thinking budget of 0, and at most 3{,}000 output tokens.

The prompt asks whether the semantic target of the specified person changes: a change of the gaze coordinates or of the mask identifier, an occlusion, a blink, or camera motion is not a shift by itself. Person identities and in-frame status are inherited from Stage~1, and the verifier cannot change them. Out-of-frame segments receive separate rules. A pattern $A\rightarrow\texttt{oof}\rightarrow A$ counts as two shifts only when the out-of-frame interval is a deliberate fixation, and as no shift when the person keeps attending to $A$ through an occlusion or a loss of visibility. A pattern $A\rightarrow\texttt{oof}\rightarrow B$ counts as one shift when the out-of-frame interval is only a pass-through, and as two shifts when it contains a complete fixation. Each candidate is resolved as a single shift, a single shift through a pass-through interval, two shifts, no shift, or unclear. Resolved shifts are kept, no-shift candidates are retained as hard negatives, and unclear candidates are sent to human review. The prompt labels an item that a person holds, wears, or manipulates as \texttt{human} and keys it to that person, for example \texttt{p001:phone}. When the window labels are built, such an item keeps this key but is treated as an \texttt{other} target, consistent with the held-object keys of Section~\ref{sec:pipeline}. Appendix~\ref{app:verifier-prompt} gives the full prompt.

\paragraph{Stage 3: human verification.}
Candidates are routed to human reviewers when mask agreement is low, the gaze displacement is small, a temporal measurement lies close to a rule threshold, the semantic target is ambiguous, or the verifier abstains or conflicts with the rule evidence. Reviewers inspect the original frames and the contact sheet, and confirm or correct the shift decision, the two endpoint targets, and the transition boundaries. Cases that remain unresolved are marked as unverified and excluded from training and evaluation. Each audit randomly samples $15\%$ of all annotations, requires a sampled error rate below $1\%$, and is followed by manual correction of the errors found. An annotation is counted as an error when the auditor would change its shift decision, either endpoint target, or a transition boundary. Audits were performed by the authors. The sampled error rate in the final audit round is $0.5\%$. On the candidates routed to human review, the shift decisions of the semantic verifier agree with those of the human reviewers in $99.4\%$ of cases.

\subsection{Verifier prompt}
\label{app:verifier-prompt}

The same prompt is used for every Stage-2 candidate on both VAT and ChildPlay. We list an abridged version below; the definitions of the individual output fields, the formatting rules for target keys, the validation of person identifiers, and the rules for rarer failure cases (shot changes, clip boundaries, and hand--object ambiguity) are omitted.

\begin{lstlisting}
You are the final event-level semantic gaze-shift auditor after Step 1, a deterministic rule-based filter. Return exactly one event-level JSON object matching the response schema.

TASK INPUT
- clip, gazer, and candidate kind
- Step 1 before / intermediate / after target keys and their immutable person ids
- transition type and before / intermediate / after dwell times
- montage tile metadata

CORE TASK
- Decide whether the gazer's real-world semantic target changed.
- A coordinate change, mask-id change, missing gaze point, OOF annotation, occlusion, blink, tracking loss, or camera movement is not by itself a gaze shift.
- Use `unclear` whenever the sampled evidence cannot reliably distinguish target continuity from a real redirection.

MONTAGE EVIDENCE
- Yellow/red gazer bbox means annotated IN/OOF; an OOF annotation is authoritative for that frame. Green shows gaze; cyan shows the SAM3 target; other masks/boxes show people and ids.
- Inspect the full sequence: gaze trajectory, head/eye direction, eyelids, target and occluder motion, and camera motion.
- SAM3 can drift, split, merge, or switch tracks; decide from gaze and semantics, not from mask ids.
- PRIOR_CONTEXT and RECOVERY tiles show surrounding targets and possible returns; they are context, not additional endpoints.

SEMANTIC CLASSES
- `human`: a person or body part, plus an item currently held, worn, carried, or manipulated by a person.
- `object`: an independent item, surface, screen, furniture, background, or region.
- `oof`: the gaze target is outside the visible image in that frame.

IMMUTABLE PERSON IDS
- Person identity has already been decided by Step 1. Copy Step 1 person ids verbatim; never perform identity recognition or invent an id.
- Preserve the Step 1 IN/OOF state and, except for person-associated items, the Step 1 human/object class.

A -> OOF -> A
- Two real shifts if the gaze forms a deliberate OOF fixation and then returns to A.
- No shift if the gazer keeps attending to A and OOF is only occlusion, blink, annotation loss, or visibility loss.

A -> OOF -> B
- One shift (pass-through) if attention moves from A to B and OOF is only transition, occlusion, or annotation loss.
- Two shifts if a complete fixation occurs during OOF before the shift to B.
- No shift if A and B are the same target under mask drift or track switching.
- Unclear if the samples cannot separate pass-through from fixation; do not decide from OOF duration alone.

EVENT OUTPUT
- `event_resolution`: one of `atomic_shift`, `pass_through_single_shift`, `two_real_shifts`, `no_shift`, or `unclear`.
- For a resolved shift, report the normalized PRE / POST (and intermediate) semantic targets, their classes, their person ids, and the resolved transitions in order.
- Give a concise reason and note any SAM3 or annotation failure.
\end{lstlisting}

% !TEX root = ../main.tex
\section{Evaluation Metrics}
\label{app:metrics}

Table~\ref{tab:metric-scope} lists the metric families and their evaluation settings. Spatial accuracy follows the official protocol of each benchmark. Unless stated otherwise, gaze-shift detection and forecasting scores are computed over the union of the VideoAttentionTarget and ChildPlay test windows; Table~\ref{tab:stage3-loss-dataset} reports them per dataset. Four-class gaze-shift AP is the macro average of one-vs-rest AP over the four classes. Binary forecasting F1 and recall use a fixed threshold of 0.5 on the shift probability for all models. Tables~\ref{tab:main-gaze-point} and~\ref{tab:main-forecast} report detection and forecasting results; this appendix also defines the temporal metrics.

\begin{table}[t]
\centering
\caption{Metric families and their evaluation scope.}
\label{tab:metric-scope}
\begin{tabular}{lccc}
\toprule
\textbf{Metric family} & \textbf{Image} & \textbf{Observed video} & \textbf{Future video} \\
\midrule
Spatial accuracy & $\checkmark$ & $\checkmark$ & $\checkmark$ \\
Temporal dynamics &  & $\checkmark$ &  \\
Gaze-shift events &  & $\checkmark$ & $\checkmark$ \\
\bottomrule
\end{tabular}
\end{table}

\subsection{Temporal metrics}
\label{app:temporal-metrics}

Frame-level L2 and AUC score each frame independently, so they cannot tell whether a prediction is stable during a fixation or whether it moves when the gaze moves. The metrics in Table~\ref{tab:ablation-prediction-temporal} measure these two properties on the prediction sequence of each test track at 8 fps. Let $\hat{g}_t$ be the arg-max of the predicted heatmap at frame $t$ and $g^*_t$ the annotated gaze point, both in normalized image coordinates. All models provide predictions on the same evaluated frames; a frame is valid when $g^*_t$ lies inside the image. Segmentation, thresholds, and aggregation are identical for every model.

\paragraph{Segments.} Segments are derived from the annotated gaze points only, so every model is evaluated on the same frames. A fixation segment is a maximal run of at least four consecutive valid frames in which $g^*_t$ moves by less than $\tau = 0.02$ between adjacent frames. A movement segment is a maximal run of at least two consecutive valid frames in which $g^*_t$ moves by at least $\tau$ between adjacent frames. Movement segments therefore include any motion of the annotated point, not only the verified gaze shifts of Appendix~\ref{app:data-engine}.

\paragraph{Jitter and Flicker.} Before computing Jitter and Flicker, we aggregate baseline predictions over the same two-frame intervals to match \modelname's temporal output resolution. Let $\mathcal{S}$ be the set of adjacent frame pairs $(t-1,t)$ within fixation segments, pooled over all tracks. Then
\begin{equation}
\text{Jitter} = \frac{1}{|\mathcal{S}|}\sum_{(t-1,t)\in\mathcal{S}} \|\hat{g}_t-\hat{g}_{t-1}\|_2,
\qquad
\text{Flicker} = \frac{1}{|\mathcal{S}|}\sum_{(t-1,t)\in\mathcal{S}} \mathbb{I}\left(\|\hat{g}_t-\hat{g}_{t-1}\|_2 > \theta\right),
\end{equation}
with $\theta = 0.10$. Jitter measures the average frame-to-frame change, and Flicker the fraction of large jumps.

\paragraph{Aligned residual.} For a movement segment $[s,e]$ in a track of $n$ frames, we compare the annotated trajectory with the prediction shifted by $k$ frames and keep the best offset within two frames (250\,ms):
\begin{equation}
E_{[s,e]}(k) = \frac{1}{|V_k|}\sum_{t\in V_k}\|\hat{g}_{t+k}-g^*_t\|_2,
\qquad
r_{[s,e]} = \min_{k\in\mathcal{K}_{[s,e]}} E_{[s,e]}(k),
\end{equation}
where $V_k = \{t \in [s,e] : 0 \le t+k < n\}$ is the set of segment frames whose shifted prediction lies inside the track, and $\mathcal{K}_{[s,e]} = \{k\in\{-2,\ldots,2\} : V_k\neq\emptyset\}$. Offsets with an empty $V_k$, which occur for short segments at the start or end of a track, are skipped; $k=0$ is always admissible, so $r_{[s,e]}$ is always defined. The reported $r$ is the mean of $r_{[s,e]}$ over all movement segments. The offset search removes short delays, so $r$ measures whether the prediction follows the path of the gaze rather than its latency. A low Jitter with a low $r$ indicates a prediction that is stable during fixations and still follows the gaze when it moves; a low Jitter with a high $r$ indicates that stability was obtained by suppressing movement.

% !TEX root = ../main.tex
\section{Additional Results}
\label{app:supp-experiments}

This section reports results that complement Section~\ref{sec:exp}. Appendix~\ref{app:image-spatial-validation} evaluates gaze-target estimation under the official per-frame protocol of each benchmark. Appendices~\ref{app:full-temporal-ablation} and~\ref{app:seed-variability} give the complete metrics of the temporal ablation and the seed variability of the latent-loss ablation. Appendix~\ref{app:latent-diagnostics} examines the future latents produced by the world predictor. Appendices~\ref{app:stage3-loss} and~\ref{app:temporal-sampling} report two design choices that were selected on the validation splits: the class weighting of the shift detection loss and the temporal sampling of the input clip. Unless stated otherwise, all experiments use GWM-L with a frozen V-JEPA~2.1 encoder, as in Section~\ref{sec:ablations}.

\subsection{Image-stage spatial validation}
\label{app:image-spatial-validation}

The GTE results in Table~\ref{tab:main-gaze-point} are computed on test windows sampled at 8\,fps, whereas prior work evaluates every annotated test frame at the native frame rate. To relate our results to this standard protocol, Table~\ref{tab:image_gte} scores every annotated frame of the official VAT and ChildPlay test sets with ground-truth head boxes. The image baselines are applied to each frame independently. GWM receives the 16-frame clip around the annotated frame, and its prediction for that frame is scored, so it uses temporal context that the image models do not have. The VLMs are prompted as in the main experiments.

GWM-g gives the lowest L2 and the highest in/out AP on both datasets. Its AUC equals that of PaGE-H+ on VAT and is 0.001 lower on ChildPlay. GWM-L is on par with the PaGE models: it matches the L2 of PaGE-H+ on VAT and has a higher in/out AP, while its AUC is 0.002 to 0.005 lower and its ChildPlay L2 is 0.001 to 0.002 higher. Gaze-LLE has an L2 more than 0.03 higher than PaGE and GWM on VAT, and the VLMs have the largest errors. The ranking agrees with that of Table~\ref{tab:main-gaze-point}: GWM is competitive with the strongest image models on per-frame localization, so its gains in shift detection do not come at the cost of spatial accuracy.

\begin{table}[t]
\centering
\caption{Image-stage gaze-target estimation on VAT and ChildPlay, evaluated per frame on the official test annotations with ground-truth head boxes. GWM rows use the final video-stage model, applied to each annotated frame with its surrounding 16-frame clip as input. This protocol differs from the 8\,fps video test windows of Tables~\ref{tab:main-gaze-point} and~\ref{tab:ablation-prediction}, so the scores are not directly comparable. On ChildPlay, Gaze-LLE is evaluated with its released VideoAttentionTarget checkpoint. AUC is not reported (--) for the VLMs, which output gaze points without a probability heatmap.}
\label{tab:image_gte}
\small
\setlength{\tabcolsep}{4pt}
\begin{tabular*}{\textwidth}{@{\extracolsep{\fill}}lcccccc@{}}
\toprule
\textbf{Model} & \multicolumn{3}{c}{\textbf{VideoAttentionTarget}} & \multicolumn{3}{c}{\textbf{ChildPlay}} \\
\cmidrule(lr){2-4} \cmidrule(lr){5-7}
 & \textbf{AUC} $\uparrow$ & \textbf{L2} $\downarrow$ & \textbf{AP} $\uparrow$ & \textbf{AUC} $\uparrow$ & \textbf{L2} $\downarrow$ & \textbf{AP} $\uparrow$ \\
\midrule
\multicolumn{7}{l}{\textit{Image gaze models}} \\
Gaze-LLE-B~\citep{ryan2025gazelle} & 0.935 & 0.107 & 0.898 & 0.956 & 0.110 & 0.997 \\
Gaze-LLE-L~\citep{ryan2025gazelle} & 0.936 & 0.103 & 0.904 & 0.959 & 0.101 & 0.995 \\
PaGE-B (distilled)~\citep{ye2026page} & \underline{0.971} & 0.069 & 0.947 & 0.970 & 0.071 & 0.997 \\
PaGE-H+~\citep{ye2026page} & \textbf{0.973} & \underline{0.066} & 0.940 & \textbf{0.972} & \underline{0.070} & 0.996 \\
\midrule
\multicolumn{7}{l}{\textit{Vision-language models}} \\
Qwen3.8-27B~\citep{qwenteam2026qwen38} & --    & 0.159 & 0.784 & --    & 0.147 & 0.979 \\
Gemini 3.8 Flash~\citep{geminiteam2026gemini38} & --    & 0.146 & 0.824 & --    & 0.119 & 0.981 \\
\midrule
GWM-B (Ours) & 0.962 & 0.069 & 0.943 & 0.963 & 0.074 & 0.995 \\
GWM-L (Ours) & 0.969 & \underline{0.066} & \underline{0.950} & 0.967 & 0.072 & \underline{0.998} \\
GWM-g (Ours) & \textbf{0.973} & \textbf{0.063} & \textbf{0.954} & \underline{0.971} & \textbf{0.069} & \textbf{0.999} \\
\bottomrule
\end{tabular*}
\end{table}

\subsection{Full metrics for the temporal ablation}
\label{app:full-temporal-ablation}

Table~\ref{tab:ablation-prediction-full} reports AUC, L2 and in/out AP on both datasets, together with the four-class shift scores, for the variants in Table~\ref{tab:ablation-prediction}; Table~\ref{tab:ablation-prediction} shows only L2. The variant $-$temporal trunk removes the two trunk blocks, so the SIM features of each tubelet reach the gaze heads without temporal aggregation. The variant $-$world predictor removes the world branch and $\mathcal{L}_{\mathrm{lat}}$. The remaining rows keep the full architecture and change only the weight $\lambda_{\mathrm{lat}}$. All variants share the image stage and the training schedule of Section~\ref{sec:implementation}.

The ChildPlay in/out AP is close to saturation for all variants (0.997 to 0.999) and does not separate them. On every other metric, the reference ($\lambda_{\mathrm{lat}}=0.1$) is best. Removing the temporal trunk causes the largest drop: VAT in/out AP decreases by 0.011, ChildPlay L2 increases by 0.017, and four-class AP decreases by 4.8 points. Above the reference weight, the effect of $\lambda_{\mathrm{lat}}$ is not monotonic: $\lambda_{\mathrm{lat}}=1.0$ is better than $\lambda_{\mathrm{lat}}=0.3$ on VAT and in the shift scores, but worse in ChildPlay AUC and L2.

\begin{table}[t]
\centering
\caption{Full observed-gaze and shift-detection metrics for the temporal modeling ablations in Table~\ref{tab:ablation-prediction}.}
\label{tab:ablation-prediction-full}

\small
\setlength{\tabcolsep}{4pt}
\renewcommand{\arraystretch}{1.16}
\definecolor{gazereference}{gray}{0.90}

{%
\begin{tabular}{@{}l*{8}{c}@{}}
\toprule
& \multicolumn{6}{c}{\textbf{Gaze-target estimation}}
& \multicolumn{2}{c}{\footnotesize\textbf{Gaze-shift detection}} \\
\cmidrule(lr){2-7}\cmidrule(lr){8-9}

\textbf{Variant}
& \multicolumn{3}{c}{\footnotesize\textbf{VideoAttentionTarget}}
& \multicolumn{3}{c}{\footnotesize\textbf{ChildPlay}}
& \multicolumn{2}{c}{\footnotesize\textbf{Four-class}} \\
\cmidrule(lr){2-4}\cmidrule(lr){5-7}\cmidrule(lr){8-9}

& {\footnotesize AUC$\uparrow$}
& {\footnotesize L2$\downarrow$}
& {\footnotesize AP$\uparrow$}
& {\footnotesize AUC$\uparrow$}
& {\footnotesize L2$\downarrow$}
& {\footnotesize AP$\uparrow$}
& {\footnotesize AP$\uparrow$}
& {\footnotesize F1$\uparrow$} \\
\midrule

\addlinespace[4pt]
$-$temporal trunk
& 0.966 & 0.073 & 0.934
& 0.958 & 0.090 & \textbf{0.999}
& 71.25 & 69.97 \\

$-$world predictor
& \underline{0.970} & 0.072 & 0.936
& \underline{0.962} & 0.087 & \underline{0.998}
& 73.92 & 73.03 \\

\rowcolor{gazereference}
$\lambda_{\mathrm{lat}}=0.1$
& \textbf{0.973} & \textbf{0.061} & \textbf{0.945}
& \textbf{0.968} & \textbf{0.073} & 0.997
& \textbf{76.09} & \textbf{75.05} \\

$\lambda_{\mathrm{lat}}=0.3$
& 0.965 & 0.067 & 0.928
& 0.958 & \underline{0.078} & \textbf{0.999}
& 74.34 & 73.65 \\

$\lambda_{\mathrm{lat}}=1.0$
& \underline{0.970} & \underline{0.064} & \underline{0.942}
& 0.955 & 0.081 & \textbf{0.999}
& \underline{74.82} & \underline{73.71} \\

\bottomrule
\end{tabular}
}
\end{table}

\subsection{Seed variability}
\label{app:seed-variability}

The ablations in Section~\ref{sec:ablations} compare single checkpoints, so we test whether the effect of $\mathcal{L}_{\mathrm{lat}}$ exceeds the variation between training runs. Table~\ref{tab:seed-variability} repeats the video-stage training of the reference model and of the variant trained without $\mathcal{L}_{\mathrm{lat}}$ (heatmap and in/out losses, world branch kept with $\lambda_{\mathrm{lat}}=0$; second row of Table~\ref{tab:ablation-loss}) with three random seeds each, keeping data, schedule and hyperparameters fixed. We report the mean and standard deviation over seeds; the main-text tables report the checkpoint selected on the validation splits. The difference in four-class AP between the two settings is 1.9~points, 2.4~times the larger seed standard deviation (Welch's $t$-test, $p=0.04$). $\mathcal{L}_{\mathrm{lat}}$ also reduces VAT L2 by 0.009 ($p=0.005$) and ChildPlay L2 by 0.012 ($p=0.007$). The difference in four-class F1 (1.1~points, $p=0.17$) has the same sign but does not reach significance. The selected reference checkpoint (76.09 four-class AP in Table~\ref{tab:ablation-loss}) lies within one standard deviation of the seed mean.
L2 in Table~\ref{tab:seed-variability} is averaged over all valid frames, as in the other tables. The L2 statistics quoted in Section~\ref{sec:ablations} instead average within each test window first and then across windows, so that every window has equal weight regardless of its number of valid frames. Under this window averaging, the ChildPlay reduction is 0.010 ($p=0.004$); the VAT reduction is 0.009 under both averages. Four-class AP and F1 are window-level scores and do not depend on this choice. In both cases the effect of $\mathcal{L}_{\mathrm{lat}}$ on L2 is significant at $p<0.01$.
% Optional: restore after running a paired bootstrap over test windows.
% A paired bootstrap over test windows (1{,}000 resamples) gives a 95\% confidence interval of [low, high]~points for the four-class AP difference between the two selected checkpoints.

% Mean followed by a smaller standard deviation, e.g. 76.09\sd{0.42}.
\newcommand{\sd}[1]{{\scriptsize$\pm$#1}}

\begin{table}[t]
\centering
\caption{Mean and standard deviation over three seeds for the latent-loss ablation (GWM-L, video stage).}
\label{tab:seed-variability}
\small
\setlength{\tabcolsep}{3pt}
\renewcommand{\arraystretch}{1.16}
\begin{tabular}{@{}lcccc@{}}
\toprule
& \multicolumn{2}{c}{\footnotesize\textbf{L2$\downarrow$}}
& \multicolumn{2}{c}{\footnotesize\textbf{Four-class}} \\
\cmidrule(lr){2-3}\cmidrule(lr){4-5}
\textbf{Setting} & {\footnotesize VAT} & {\footnotesize ChildPlay}
& {\footnotesize AP$\uparrow$} & {\footnotesize F1$\uparrow$} \\
\midrule
w/o $\mathcal{L}_{\mathrm{lat}}$
  & 0.071\sd{0.002} & 0.084\sd{0.003}
  & 73.4\sd{0.8} & 73.2\sd{0.9} \\
$\lambda_{\mathrm{lat}}=0.1$ (reference)
  & 0.062\sd{0.002} & 0.072\sd{0.002}
  & 75.3\sd{0.8} & 74.3\sd{0.7} \\
\bottomrule
\end{tabular}
\end{table}

\subsection{Latent prediction diagnostics}
\label{app:latent-diagnostics}

Table~\ref{tab:latent-diagnostics} tests whether the world predictor forecasts future scene features beyond simple copies of the observation. On the test split, we compare its future latents with the targets $z_{t,p}$ of $\mathcal{L}_{\mathrm{lat}}$, and do the same for two reference predictors: the last observed tubelet, $\mathrm{LN}(e_{7,p})$, repeated over the future positions, and the normalized mean of the observed features. Motion patches are the 20\% of future patches with the largest motion weight in $\mathcal{L}_{\mathrm{lat}}$. Both reference predictors use only observed features, so their errors measure how much of the future latent a static scene already explains.

The world predictor has the lowest error in both columns. Relative to the observed mean, it reduces the latent L1 by 3.5\% over all patches and by 5.8\% on motion patches. Because the prediction is a residual on the normalized observation mean (Section~\ref{sec:architecture}), this reduction comes from the learned residual. The larger gain on motion patches shows that the predictor models how the scene changes, not only its static content. Repeating the last observed tubelet is worse than the observed mean, which suggests that averaging over the observation reduces noise in the target estimate.

\begin{table}[t]
\centering
\caption{Future latent prediction error against the frozen-encoder targets on the test split (GWM-L).}
\label{tab:latent-diagnostics}
\small
{%
\begin{tabular}{lcc}
\toprule
\textbf{Predictor} & \textbf{Latent L1 (all)} $\downarrow$ & \textbf{Latent L1 (motion patches)} $\downarrow$ \\
\midrule
Repeat last observed tubelet & 0.460 & 0.753 \\
Observed mean                & \underline{0.452} & \underline{0.721} \\
World predictor (ours)       & \textbf{0.436} & \textbf{0.679} \\
\bottomrule
\end{tabular}%
}
\end{table}

\subsection{Class weighting of the shift detection loss}
\label{app:stage3-loss}

The four window classes of shift detection are imbalanced, so the class weights of the cross-entropy loss affect which errors the head trades off. We compare four class-weighting strategies. The shift detection head is trained on the frozen outputs of the video stage (Section~\ref{sec:implementation}), so only this head is retrained for each strategy, and all other settings are fixed. Table~\ref{tab:stage3-loss-dataset}
reports results separately for each dataset. Uniform weighting gives every class a weight of 1. Balanced weighting keeps the no-shift weight at 1 and balances the three shift types among themselves, with each weight capped at 2; inverse-frequency weighting gives all four classes equal total weight; focal loss uses $\gamma = 2$ \citep{lin2017focal}. Balanced weighting leads every reported
metric on both datasets. Uniform weighting ranks second on VideoAttentionTarget in
accuracy, F1, and AP, while inverse-frequency weighting ranks second on ChildPlay in
accuracy, F1, and M-F1 but has the lowest AP. Thus, balanced weighting gives the most
consistent performance across datasets and metrics in this comparison. Compared with inverse-frequency weighting, the cap of 2 keeps the rarest transition type from dominating the loss, and the fixed no-shift weight keeps the head from over-predicting shifts.

\begin{table}[t]
\centering
\caption{Class weighting of the four-class shift detection loss on each dataset. Acc.\ and M-F1 are four-class accuracy and macro-F1; F1 and AP are binary shift scores. All scores are percentages.}
\label{tab:stage3-loss-dataset}
\small
{%
\begin{tabular}{lcccccccc}
\toprule
\textbf{Weighting} & \multicolumn{4}{c}{\textbf{VideoAttentionTarget}} & \multicolumn{4}{c}{\textbf{ChildPlay}} \\
\cmidrule(lr){2-5} \cmidrule(lr){6-9}
 & Acc.$\uparrow$ & F1$\uparrow$ & M-F1$\uparrow$ & AP$\uparrow$ & Acc.$\uparrow$ & F1$\uparrow$ & M-F1$\uparrow$ & AP$\uparrow$ \\
\midrule
Balanced          & \textbf{81.80} & \textbf{81.20} & \textbf{74.95} & \textbf{87.33} & \textbf{77.82} & \textbf{84.75} & \textbf{79.26} & \textbf{89.99} \\
Uniform           & \underline{80.94}          & \underline{79.85}          & 71.80          & \underline{87.12}          & 73.52          & 78.67          & 75.14          & \underline{89.53}          \\
Inverse-frequency & 75.89          & 77.27          & 69.76          & 86.85          & \underline{76.90}          & \underline{82.55}          & \underline{75.60}          & 86.65          \\
Focal ($\gamma = 2$) & 78.80       & 78.59          & \underline{72.68}          & 86.08          & 75.39          & 77.41          & 74.93          & 88.00          \\
\bottomrule
\end{tabular}%
}
\end{table}

\subsection{Temporal sampling and learning rate}
\label{app:temporal-sampling}

Each sampling setting trains and evaluates the video stage with a different input clip, defined by its number of frames and its frame rate. Together they set the temporal span of the observation: 8 frames at 8\,fps and 16 frames at 16\,fps cover 1\,s; 8 frames at 4\,fps, 16 frames at 8\,fps (the main setting) and 32 frames at 16\,fps cover 2\,s; 16 frames at 4\,fps and 32 frames at 8\,fps cover 4\,s. The learning-rate variant keeps the main sampling and halves the peak learning rate. Table~\ref{tab:temporal-sampling} reports test results for these settings; the reference was selected on the validation splits. The reference ($16$ frames at $8$ fps) has the lowest L2 and highest AUC on both datasets. Among the other sampling settings, $32$ frames at $8$ fps gives the lowest L2 (tied on VideoAttentionTarget with $16$ frames at $16$ fps); frame-rate effects vary across datasets. Halving the learning rate gives worse scores than the reference on both datasets. The three settings that cover 2\,s differ in VAT L2 by up to 0.016, so the result depends on how densely the span is sampled and not only on its length. A longer span does not help either: 32 frames at 8\,fps covers twice the span of the main setting but has a higher L2 on both datasets. The sampling settings use different valid-window sets, so the comparisons are not strictly paired.

\begin{table}[t]
\centering
\caption{Temporal sampling and learning-rate sensitivity for gaze-target estimation.}
\label{tab:temporal-sampling}
\begin{tabular}{lcccc}
\toprule
& \multicolumn{2}{c}{\textbf{VideoAttentionTarget}} & \multicolumn{2}{c}{\textbf{ChildPlay}} \\
\cmidrule(lr){2-3} \cmidrule(lr){4-5}
\textbf{Setting} & L2$\downarrow$ & AUC$\uparrow$ & L2$\downarrow$ & AUC$\uparrow$ \\
\midrule
$8$f @ $4$ fps & 0.077 & 0.962 & 0.092 & \underline{0.964} \\
$8$f @ $8$ fps & 0.072 & 0.965 & 0.089 & 0.963 \\
$16$f @ $4$ fps & 0.070 & 0.964 & 0.089 & 0.954 \\
$16$f @ $16$ fps & 0.069 & \underline{0.970} & 0.088 & 0.958 \\
$32$f @ $8$ fps & 0.069 & 0.966 & \underline{0.084} & 0.959 \\
$32$f @ $16$ fps & 0.073 & \underline{0.970} & 0.087 & 0.956 \\
\ours \textbf{Main ($16$f @ $8$ fps)} & \textbf{0.061} & \textbf{0.973} & \textbf{0.073} & \textbf{0.968} \\
Learning rate $\times 0.5$ & \underline{0.068} & 0.969 & 0.089 & 0.959 \\
\bottomrule
\end{tabular}
\end{table}

\section{Additional Qualitative Examples}
\label{app:qual}
\begin{figure}[p]
    \centering
    \includegraphics[width=0.9\textwidth]{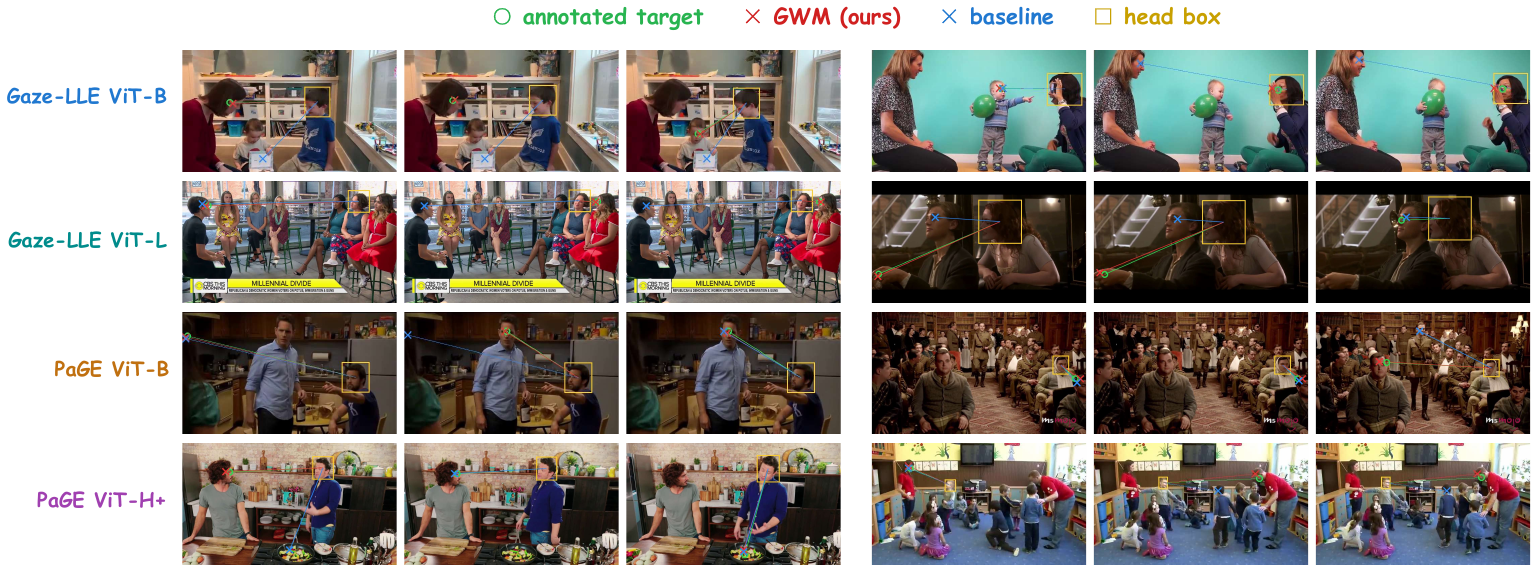}
    \caption{Gaze-target estimation on consecutive observed frames. Each row compares \modelname (red crosses) with one baseline (blue crosses); green circles mark annotated targets and yellow boxes mark the specified people.}
    \label{fig:qual-gte}

    \vspace{6pt}
    \includegraphics[width=0.8\textwidth]{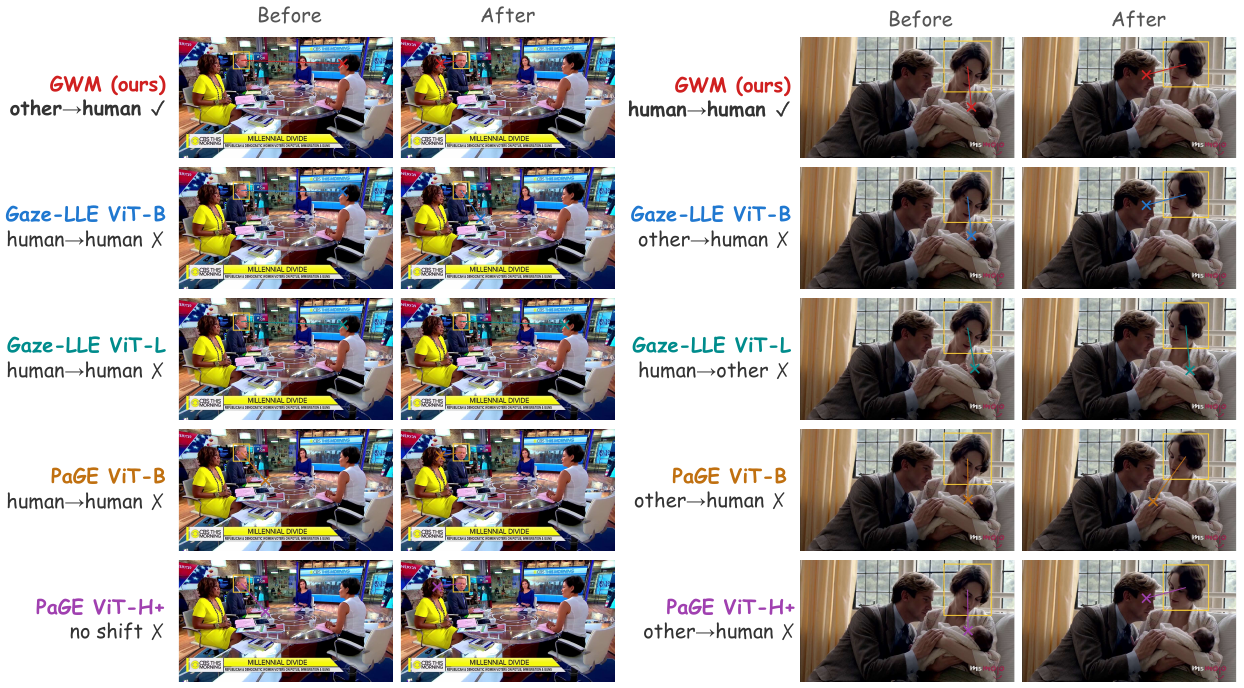}
    \caption{Gaze-shift detection in two observed clips. Each row shows the frames before and after the annotated transition, together with the transition type predicted by each model ($\checkmark$ correct, $\times$ incorrect).}
    \label{fig:qual-shift-detection}

    \vspace{6pt}
    \includegraphics[width=0.95\textwidth]{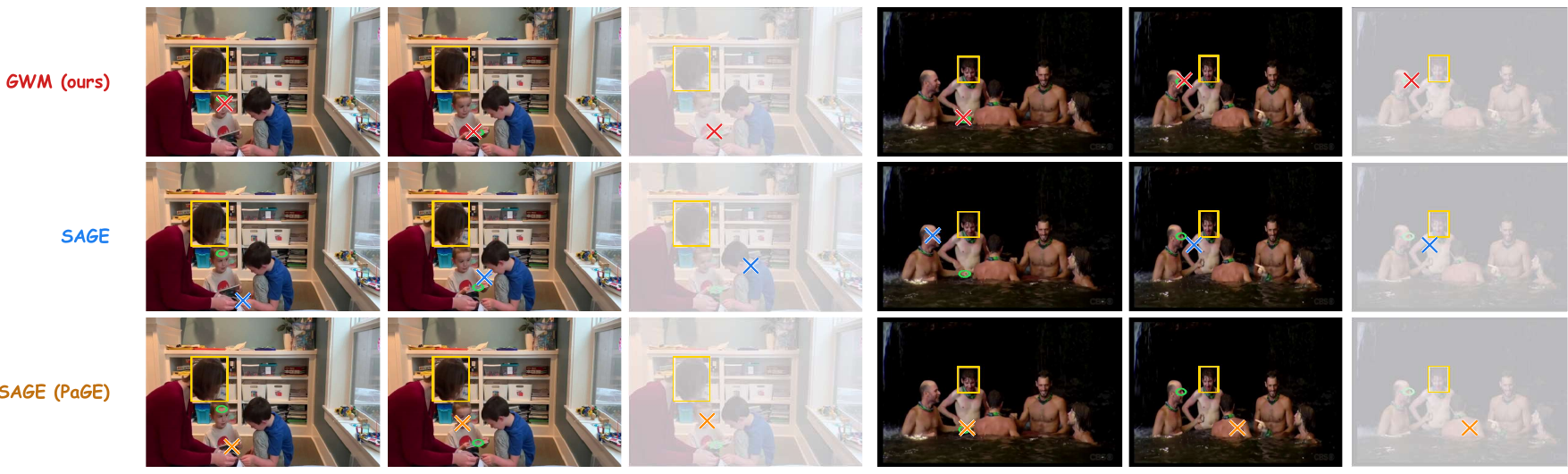}
    \caption{Gaze-target forecasting compared with SAGE with Sharingan and SAGE with PaGE. Each three-column example shows an observed frame, a future frame at the shift onset, and a later future frame (faded). Green circles mark annotated gaze targets; crosses show predictions from \modelname (red), SAGE with Sharingan (blue), and SAGE with PaGE (orange). Yellow boxes mark the specified people. Future frames are shown for reference only and are not model inputs.}
    \label{fig:qual-forecasting}
\end{figure}

Figures~\ref{fig:qual-gte}--\ref{fig:qual-forecasting} compare \modelname with external baselines on observed gaze-target estimation, gaze-shift detection, and future gaze-target forecasting. The localization examples show predictions across consecutive frames in scenes with multiple people and competing targets. The detection examples show other-to-human and human-to-human transitions; baseline errors include confusing transition types and predicting no shift. The forecasting examples compare \modelname with SAGE with Sharingan and SAGE with PaGE on the same clips, showing how predicted gaze locations track changes in the annotated target beyond the observation window. These selected cases illustrate model behavior in specific scenes and complement the quantitative evaluation.

% The three qualitative figures share one float page so that no page is left mostly empty.

\clearpage
% !TEX root = ../main.tex
\section{Limitations and Future Work}
\label{app:limitations}

\paragraph{Temporal scope.}
The present experiments use finite observation windows and fixed forecast horizons, with gaze outputs produced at two-frame tubelet resolution. Future work could extend the model to longer context, multiple forecast horizons, and finer temporal readout, and evaluate event-boundary localization alongside window-level detection and forecasting.

\paragraph{Subject conditioning and target representation.}
An important next step is to evaluate subject conditioning with automatically detected and temporally tracked head boxes, including sensitivity to crop quality. Adding explicit target representations and temporal object association could support instance-level gaze transitions, distinguish objects of the same category, and connect latent future predictions to interpretable target identities. Richer treatment of off-screen targets would further broaden the event vocabulary.

\paragraph{Annotation and evaluation.}
Further work could expand the human audit across datasets and interaction settings, quantify agreement among independent human annotators, and evaluate additional transition types under a common protocol. Such extensions would complement the current person-centered detection taxonomy and binary forecasting task.

\end{document}